\documentclass[10pt,twocolumn,letterpaper]{article}

\usepackage{iccv}
\usepackage{times}
\usepackage{epsfig}
\usepackage{graphicx}
\usepackage{graphics}
\usepackage{amsmath,bm}
\usepackage{amssymb}
\usepackage{subcaption}
\usepackage{booktabs}

\usepackage{bbm}

\usepackage{caption}
\usepackage{subcaption}
\usepackage{xspace}
\usepackage{wrapfig}
\usepackage[ruled,linesnumbered]{algorithm2e}
\usepackage{multirow}

\newcommand{\methodname}{BGS\xspace}
\newcommand{\methodnamefull}{Group-class \textbf{B}alanced \textbf{G}reedy \textbf{S}ampling\xspace}
\newcommand{\ours}{\methodname}

\usepackage[pagebackref=true,breaklinks=true,letterpaper=true,colorlinks,bookmarks=false]{hyperref}
\usepackage{cleveref}
\iccvfinalcopy 

\def\iccvPaperID{9134} 

\ificcvfinal\fi

\begin{document}

\title{Continual Learning in the Presence of Spurious Correlation}

\author{Donggyu Lee\textsuperscript{\rm 1}\thanks{Equal contribution.},\ \ Sangwon Jung\textsuperscript{\rm 2}\footnotemark[1],\ \ Taesup Moon\textsuperscript{\rm 2,3,4}\footnote{Corresponding author.}\\
\textsuperscript{\rm 1} Department of Electrical and Computer Engineering, Sungkyunkwan University\\\
\textsuperscript{\rm 2} Department of Electrical and Computer Engineering, Seoul National University\\\
\textsuperscript{\rm 3} SNU-LG AI Research Center \ \ 
\textsuperscript{\rm 4} ASRI/INMC/IPAI/AIIS, Seoul National University \\
{\tt\small ldk3088@skku.edu, \ \{s.jung,tsmoon\}@snu.ac.kr}}

\maketitle
\ificcvfinal\thispagestyle{empty}\fi

\begin{abstract}

Most continual learning (CL) algorithms have focused on tackling the stability-plasticity dilemma, that is, the challenge of preventing the forgetting of previous tasks while learning new ones. However, they have overlooked the impact of the knowledge transfer when the dataset in a certain task is \textit{biased} --- namely, when some unintended spurious correlations of the tasks are learned from the biased dataset. In that case, how would they affect learning future tasks or the knowledge already learned from the past tasks? In this work, we carefully design systematic experiments using one synthetic and two real-world datasets to answer the question from our empirical findings. Specifically, we first show through two-task CL experiments that standard CL methods, which are unaware of dataset bias, can transfer biases from one task to another, both forward and backward, and this transfer is exacerbated depending on whether the CL methods focus on the stability or the plasticity. We then present that the bias transfer also exists and even accumulate in longer sequences of tasks. Finally, we propose a simple, yet strong plug-in method for debiasing-aware continual learning, dubbed as \methodnamefull (\ours). As a result, we show that our \ours can always reduce the bias of a CL model, with a slight loss of CL performance at most. 
\end{abstract}

\section{Introduction}
Continual learning (CL) is essential for a system that needs to learn (potentially increasing number of) tasks from sequentially arriving data. The main challenge of CL is to overcome the \textit{stability-plasticity} dilemma \cite{mermillod2013stability}; when a CL model focuses too much on the stability, it would suffer from low plasticity for learning a new task (and vice versa). Recent deep neural networks (DNNs) based CL methods \cite{ewc, agscl, li2017learning} attempted to address the dilemma by devising mechanisms to attain stability while improving plasticity thanks to the \textit{knowledge transferability} \cite{tan2018survey}, which is one of the standout properties of DNNs. Namely, while maintaining the learned knowledge, the performance on a new task (resp. past tasks) is improved by transferring knowledge of past tasks (resp. a new task). Such phenomena are called the forward and backward transfer, respectively.

\begin{figure}[t!]
    \centering
    \includegraphics[width=0.8\linewidth]{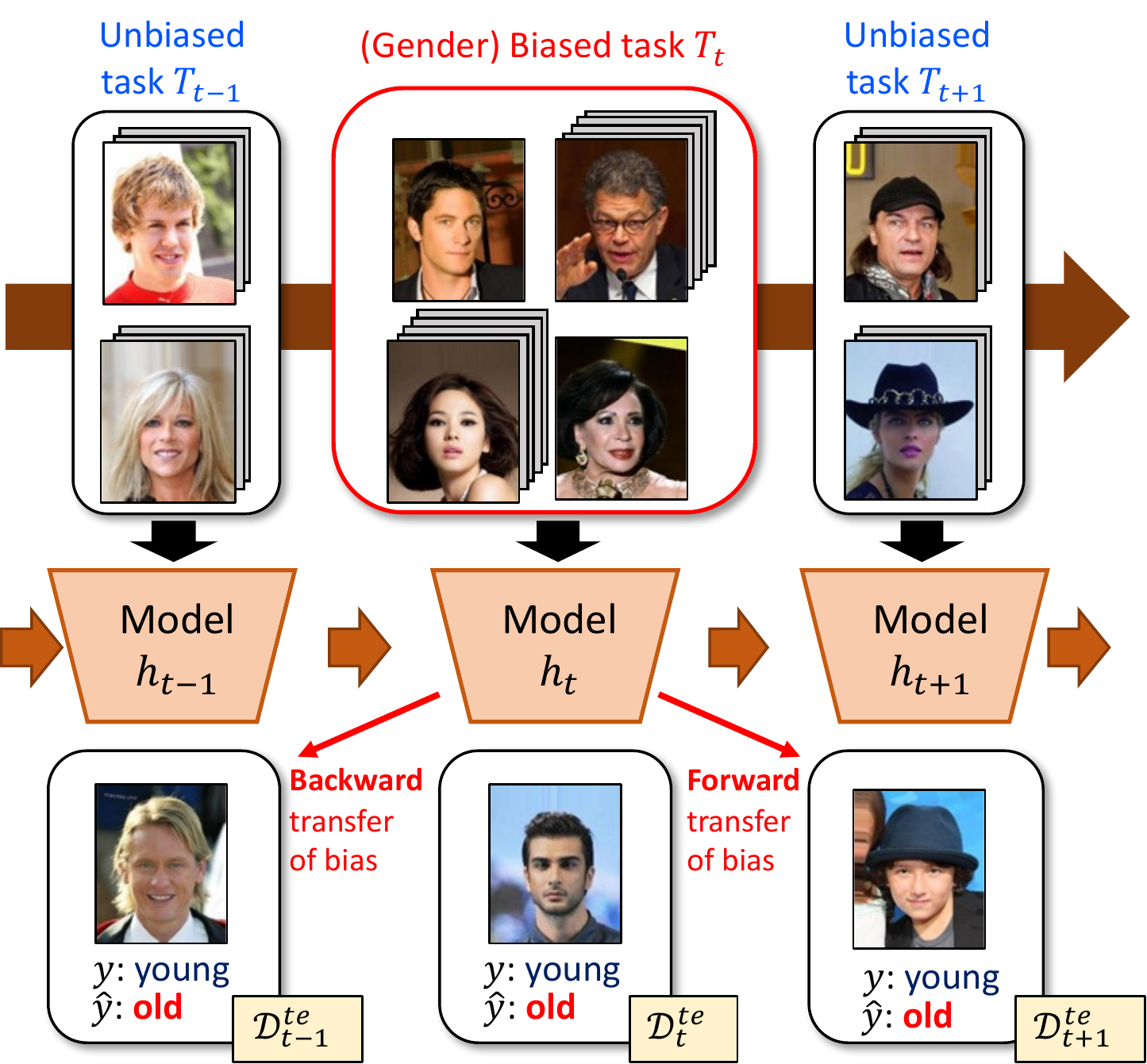}
    \caption{\small{\bf An illustration for bias transfer in continual learning.} The three tasks sequentially arrive,  where each sample contains blond hair, black hair and hat. When a naive CL model is used to update the model $h_t$ from $T_t$, the gender bias is obtained by the model, and the bias is transferred forward and backward, \eg, the model falsely predicts young man images to ``old'' class in $T_{t-1}$ and $T_{t+1}$.} 
    \label{fig:intro_figure}
    \vspace{-.2in}
\end{figure}

While such DNN-based approaches for CL have been successful to some extent, they have not explicitly considered a more realistic and challenging setting in which the \textit{dataset bias} \cite{torralba2011unbiased} exists; \textit{i.e.}, a training dataset may contain unintended correlations between some spurious features and class labels. 
In such a case, it is widely known that DNNs often dramatically fail to generalize to the test data without the correlation due to learning the spurious features \cite{groupdro, bahng2020learning}.
For instance, consider a DNN that can accurately classify birds in the sky. However, when presented with images of birds outside of their typical sky background, the model may fail due to relying on shortcut strategies that exploit the background \cite{geirhos2020shortcut}.
This issue has been the subject of various attempts to resolve it, with earlier approaches \cite{learningfromfailure, liu2021just} often being based on empirical observations of DNN behavior, which yielded suboptimal results.


Now, we claim that the issue of learning spurious correlations in the context of CL can be a significant problem because it can lead to the issue of \textit{bias transfer}. In a recent study \cite{salman2022does}, it is shown that the bias learned by a model can be (forward) transferred to the downstream model even when it is fine-tuned with unbiased downstream task data. 
In CL, this issue can be potentially exacerbated since CL involves learning a sequence of tasks, and the bias transfer can occur in both forward and backward directions. 
For instance, consider an example of domain-incremental learning (Domain-IL) setting shown in Figure \ref{fig:intro_figure}, in which the goal is to incrementally learn the classifier that predicts whether the face in the image is \textit{young} or not, as the training data arrives.  
Now, assume that among three tasks, $T_{t-1}$, $T_{t}$ and $T_{t+1}$, only the dataset for $T_t$ possesses the \textit{gender} bias due to the data imbalance; namely, ``male'' and ``female'' face tends to spuriously correlate with ``old'' and ``young'' class, respectively. 
We argue that when a naive CL method, which does not concern about the bias transfer, is used to update the model $h_t$ from $T_t$, the \textit{gender} bias picked up by the model can adversely affect the prediction for the test images in previous or future task, \textit{i.e.}, $T_{t-1}$ or $T_{t+1}$. 
In other words, while the models independently trained with $T_{t-1}$ or $T_{t+1}$ would not contain any bias, the naive CL-updated $h_{t}$ and $h_{t+1}$ can falsely predict the test ``male'' images for $T_{t-1}$ or $T_{t+1}$ to be ``old'', respectively, due to the ``gender'' bias in $h_t$ transferring to the predictions for past and future tasks. To the best of our knowledge, this issue has not been carefully investigated in the CL research community. 



To that end, through systematic and carefully designed experiments, we quantitatively show that the above issue of bias transfer in CL, both forward and backward, indeed exists and significantly affects the model performance. More specifically, using one synthetic and two real-world datasets, we first carry out extensive two-task CL experiments and identify that when a typical CL method focuses on the stability (resp. plasticity), the bias learned from the past task (resp. current task) gets transferred and affects the model learned for the current task (resp. past task). Furthermore, we show such forward and backward bias transfer also exists and even \textit{accumulates} when naively applying CL methods for a longer sequence of tasks with dataset bias. Finally, we present a simple yet strong plug-in method, dubbed as \methodnamefull (\ours), which can be easily combined with any existing CL methods. Using an class-group balanced exemplar memory, \ours retrains the last layer of a DNN-based CL model after learning the last task. As a result, we show that our method can always reduce the bias of a CL model with a slight loss of accuracy at most.
Despite of our improvements, our results clearly call for a fundamental and novel approach for continually learning each task with potential bias while debiasing the task. 

The remainder of this paper is organized as follows. Section \ref{sec:relwork} discusses some related works for continual and debiasing learning, and \ref{sec:setup} introduces the experimental setup. In Section \ref{sec:two_task_studies}, we first investigate the forward and backward transfers of bias in two-task CL scenarios, and in Section \ref{sec:longer}, further study them in longer sequences of tasks. Then, Section \ref{sec:comparison} gives a detailed description of \ours and comparison results with other baseline methods.

\section{Related works}
\label{sec:relwork}

\noindent\textbf{Continual learning (CL)}.
Assuming that tasks are clearly separated, CL scenarios are typically categorized into three categories: task-incremental learning (Task-IL), domain-incremental learning (Domain-IL), and class-incremental learning (Class-IL) \cite{van2019three}. The Task-IL and Class-IL assume each task has a disjoint set of labels and the task identity is provided during training. The main difference between the two is whether the task identity is used (Task-IL) at test time or not (Class-IL). Task-IL adopts the structure of a multi-headed network with a classification head for each task since it uses the task identity at the test time. In contrast, Class-IL uses a single-headed network due to the lack of task identity at the test time. In Domain-IL, the class set of the tasks always remains the same, but only the input distributions vary as the number of tasks increases.  

Recent CL methods can also be classified into three categories based on how they prevent forgetting of the previously learned tasks \cite{de2021continual}: rehearsal based, regularization based, and parameter isolation based methods. Regularization based methods add regularization terms for penalizing deviation from past models and balance the stability-plasticity trade-off by controlling the regularization hyperparameter \cite{ewc, li2017learning, agscl, titsias2020functional}. Rehearsal based methods store some data points from past tasks in a small exemplar memory and replay them while learning the current task \cite{er, gem, agem, tiwari2022gcr}. Parameter isolation based methods \cite{rusu2016progressive, mallya2018packnet, mallya2018piggyback, hung2019compacting, ye2021lifelong} allocate model parameters separately for each task by masking out previous task parameters and updating only remained parameters for learning a new task. Note that parameter isolation based methods can be applied to only Task-IL settings since they require task identity during inference to separate parameters for each task.

While many state-of-the-art CL methods in each category have been actively developed, many of them were limited for direct application to real-world deployment scenarios. 
For instance, most existing CL methods focus on learning well-balanced tasks with reliable labels. To that end, some considered more practical settings; \textit{e.g., } CL scenarios with imbalanced data \cite{kim2020imbalanced}, noisy labels \cite{kim2021continual, bang2022online}, and dataset bias \cite{lesort2022continual}. While \cite{lesort2022continual} shares a somewhat similar motivation as ours, their overall experimental setup was not sufficiently fine-grained to fully support their own findings.


\noindent\textbf{Spurious correlations and debiased learning}.  
 Machine learning community has attempted to comprehend issues of spurious correlations, and as a result, there have been a variety of studies that aim to identify and mitigate various forms of real-world spurious correlations. For instance, in vision tasks, neural networks may rely on background \cite{groupdro, geirhos2020shortcut}, texture \cite{geirhos2018}, or semantically irrelevant features of objects \cite{liu2022contextual} in the image. In addition, numerous approaches to address these issues have been proposed. 
Early approaches utilize known group labels that indicate the misleadingly correlated attributes, such as background or gender \cite{groupdro, mfd}. As an example, Sagawa et al. \cite{groupdro} developed Group DRO which minimizes the worst-case group loss in the training data by using some prior knowledge of spurious correlations. Since annotating group labels is costly, more recent works have developed more practical methods that need only bias type \cite{bahng2020learning}, partially annotated group labels \cite{liu2021just, jung2022learning}, or no group labels  \cite{learningfromfailure}. Nonetheless, they have not been considered spurious correlations in continual learning.

\section{Experimental setup}
\label{sec:setup}

\subsection{Notations and problem setting}
We consider the \textit{bias-aware continual learning} which is composed of a sequence of classification tasks which may contain a dataset bias. Each data sample in the $t$-th task $T_t$ consists of the tuple $(x_i,a_i,y_i)$ where $a_i\in\mathcal{A}$ and $y_i\in\mathcal{Y}_t$ are group and class labels of an input $x_i$, respectively. 
Unless otherwise noted, we consider the single type of bias in a CL scenario for simplicity of analysis. In addition, the sets of class label, $\{\mathcal{Y}_i\}_{i\in\{ 1, 2, \dots \}}$, can be identical or not, depending on the type of CL scenarios. 

\subsection{Benchmark datasets}
We use one synthetic dataset, Split CIFAR-100S, and two real-world datasets, CelebA$^2$ (or CelebA$^8$) and Split ImageNet-100, which are applied for Task-IL, Domain-IL, and Class-IL settings, respectively. Followings are descriptions for our datasets (more details are in Appendix). 

\noindent\textbf{Split CIFAR-100S} is a modified dataset from Split CIFAR-100 \cite{si, er, vandeven2020brain}, which randomly divides CIFAR-100 \cite{cifar10} into 10 tasks with 10 distinct classes. Similarly as in \cite{wang2020towards}, we modify split CIFAR-100 such that, given a skew-ratio $\alpha\geq 0.5$, half of the classes in each task are skewed toward the grayscale group and the other half toward the color group; namely, the training images of each class are split into $\alpha$ and $1-\alpha$ ratios for each group. Thus, the ``color'' becomes the bias of the dataset. We set 7 bias levels (0-6) by dividing the range of skew-ratio from 0.5 to 0.99 evenly on a log scale for systematic control of the degree of bias. 

\noindent\textbf{CelebA} \cite{liu2015deep} contains more than 200K face images, each annotated with 40 binary attributes. It is notorious for containing representation biases towards specific attributes such as race, age, or gender \cite{torfason2017face,fabbrizzi2022survey}; for instance, the sub-populations of young women and old men are over-represented in the CelebA dataset. Unless otherwise specified, we use ``gender'' and ``young'' attributes as a group and a class label, respectively. We additionally select one or three other attributes and based on them, divide CelebA into two tasks (used in \cref{sec:two_task_studies}) or eight tasks (used in \cref{sec:longer} and \ref{sec:comparison}), which are denoted as CelebA$^2$ or CelebA$^8$, respectively. Since the representation bias in CelebA can be controlled by adjusting imbalances between the class and group labels, we set 7 bias levels varying on the degree of imbalance from 0.5 to 0.99, as the same procedure in Split CIFAR-100S.

\noindent\textbf{Split ImageNet-100} divides ImageNet-100 \cite{(Imagenet)Deng09} into 10 tasks with disjoint 10 classes which are randomly sampled from original 1000 classes in ImageNet-1000. 
It is known that DNN models trained from ImageNet are biased towards watermark \cite{singla2021salient}; namely, the ImageNet-pretrained models predict an image as the carton class when injecting a watermark to it, since most of carton images in the ImageNet training dataset include the watermark. 
Following the recent study, we also consider the watermark bias in Split ImageNet-100.
We utilize two types of test datasets, the original ImageNet validation dataset, and ImageNet-W with watermarks injected by style transfer \cite{li2022whac, gatys2016image}, in order to measure the degree of bias. We set a bias level of a task as 0 or 6, depending on the presence of a carton class in the task.

\begin{figure*}[t!]
    \centering
    \begin{subfigure}[t]{0.3\linewidth}
        \includegraphics[width=0.9\linewidth]{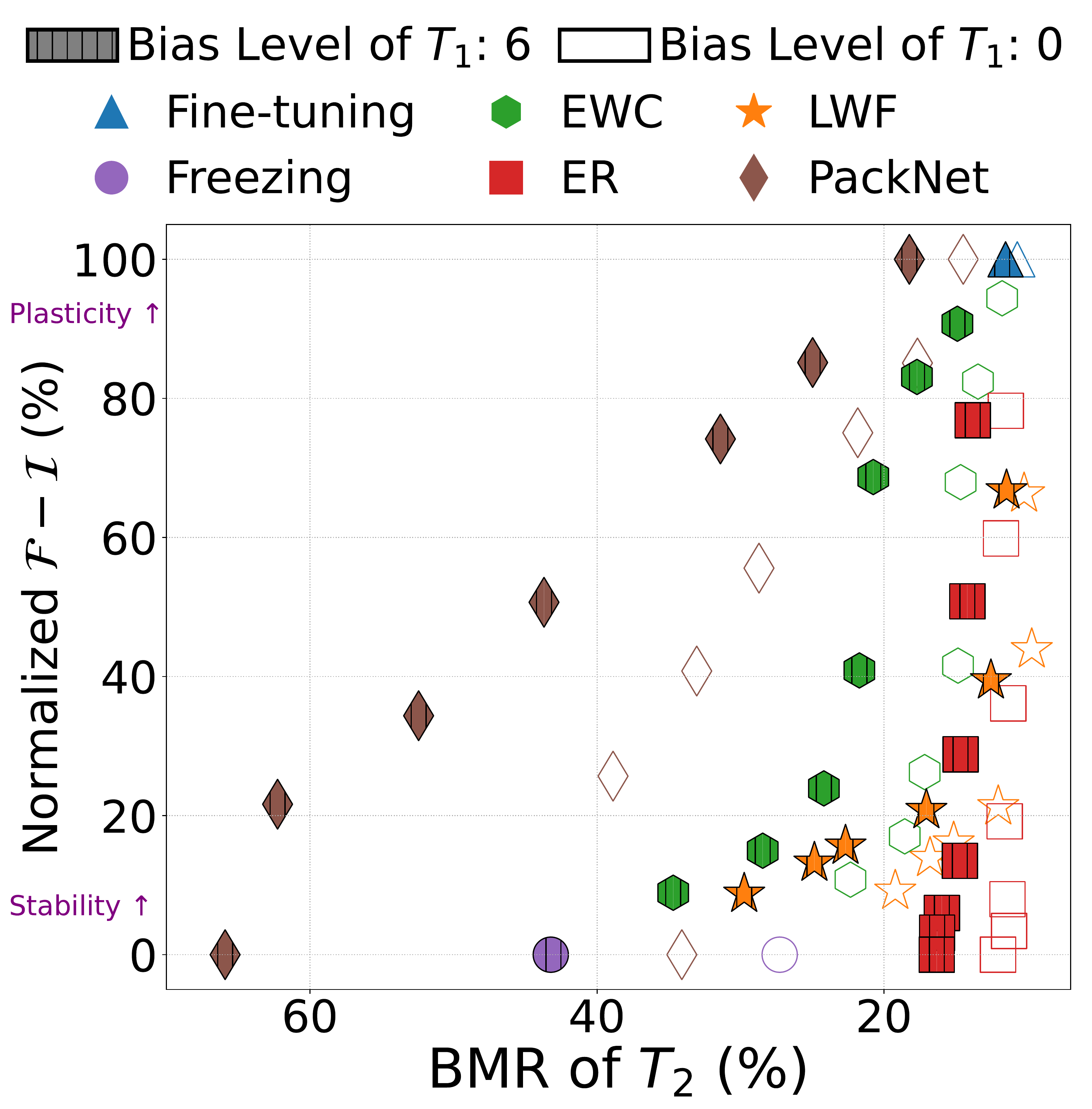}
        \caption{\small{Split CIFAR-100S in Task-IL}}
        \label{fig:two_forward_cifar}
    \end{subfigure}
    \begin{subfigure}[t]{0.3\linewidth}
        \includegraphics[width=0.9\linewidth]{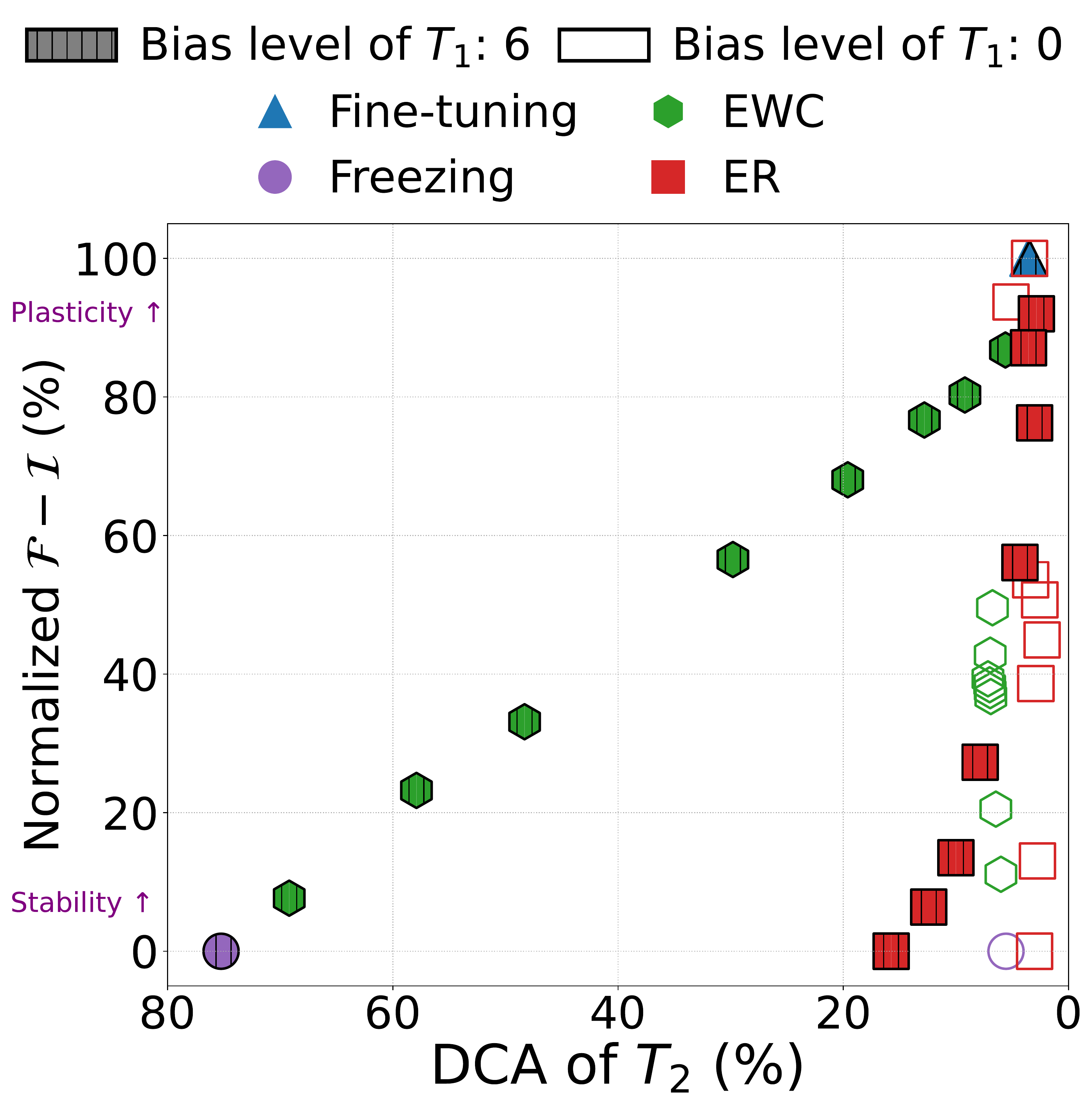}
        \caption{CelebA$^2$ in Domain-IL}
        \label{fig:two_forward_celeba}
    \end{subfigure}
    \begin{subfigure}[t]{0.3\linewidth}
        \includegraphics[width=0.9\linewidth]{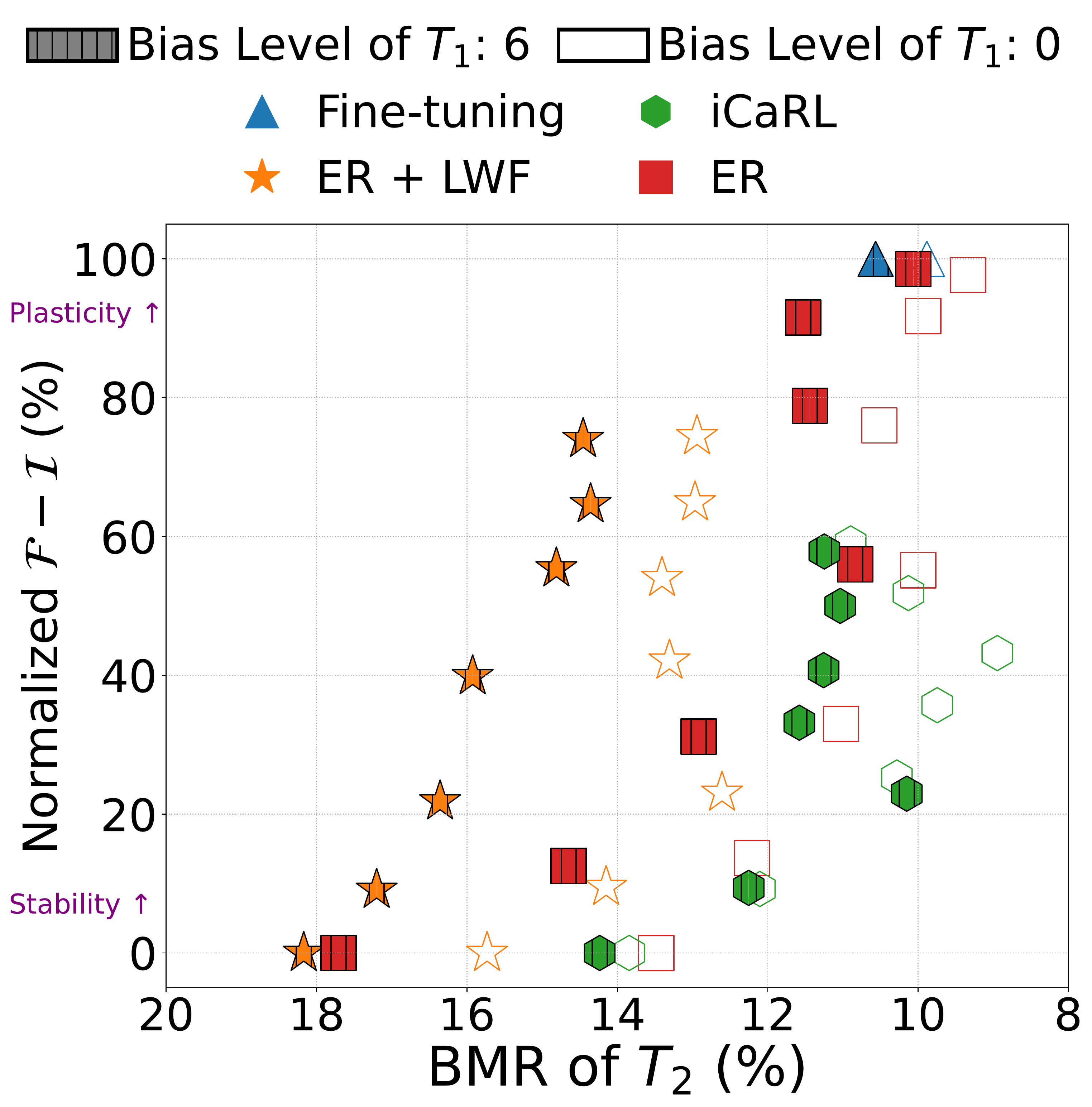}
        \caption{Split ImageNet-100 in Class-IL}
        \label{fig:two_forward_imagenet}
    \end{subfigure}
    \vspace{-.05in}
    \caption{\small {\bf Forward transfer of bias in two tasks-continual learning.}}
    \label{fig:two_forward}
\end{figure*}

\begin{figure*}[t!]
    \centering
    \begin{subfigure}[t]{0.3\linewidth}
        \includegraphics[width=0.9\linewidth]{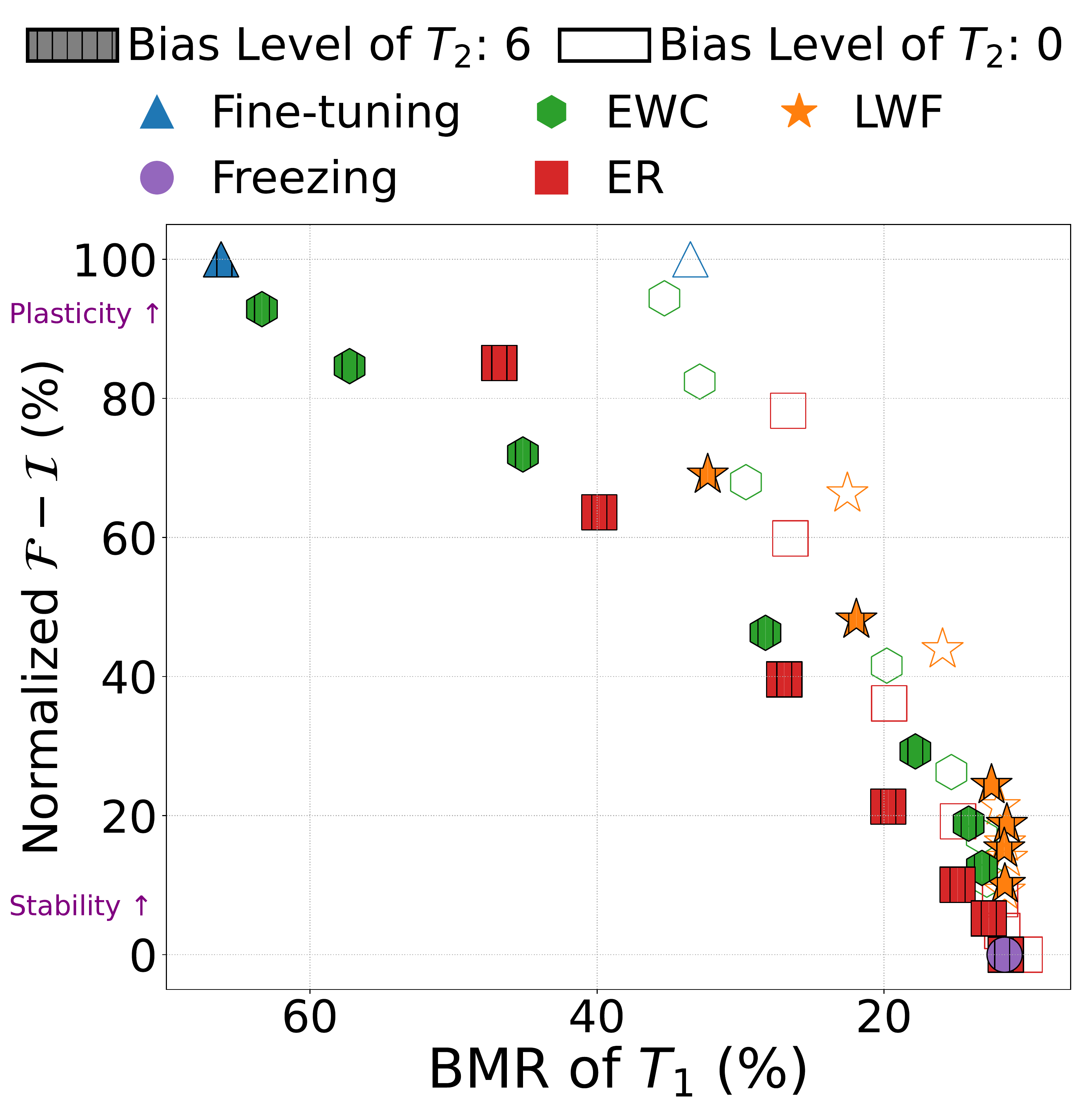}
        \caption{Split CIFAR-100S in Task-IL}
        \label{fig:two_backward_cifar}
    \end{subfigure}
    \begin{subfigure}[t]{0.3\linewidth}
        \includegraphics[width=0.9\linewidth]{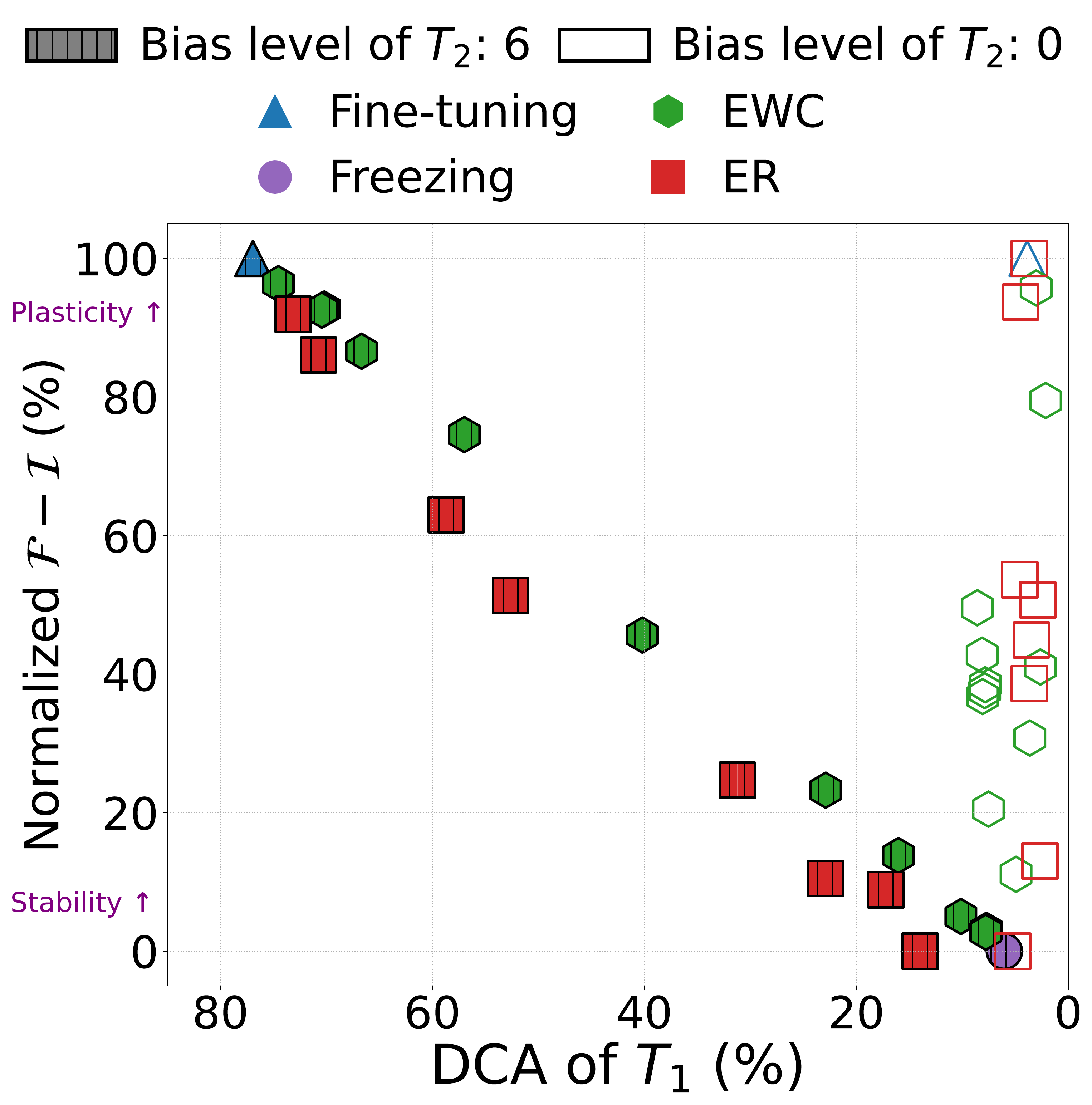}
        \caption{CelebA$^2$ in Domain-IL}
        \label{fig:two_backward_celeba}
    \end{subfigure}
    \begin{subfigure}[t]{0.3\linewidth}
        \includegraphics[width=0.9\linewidth]{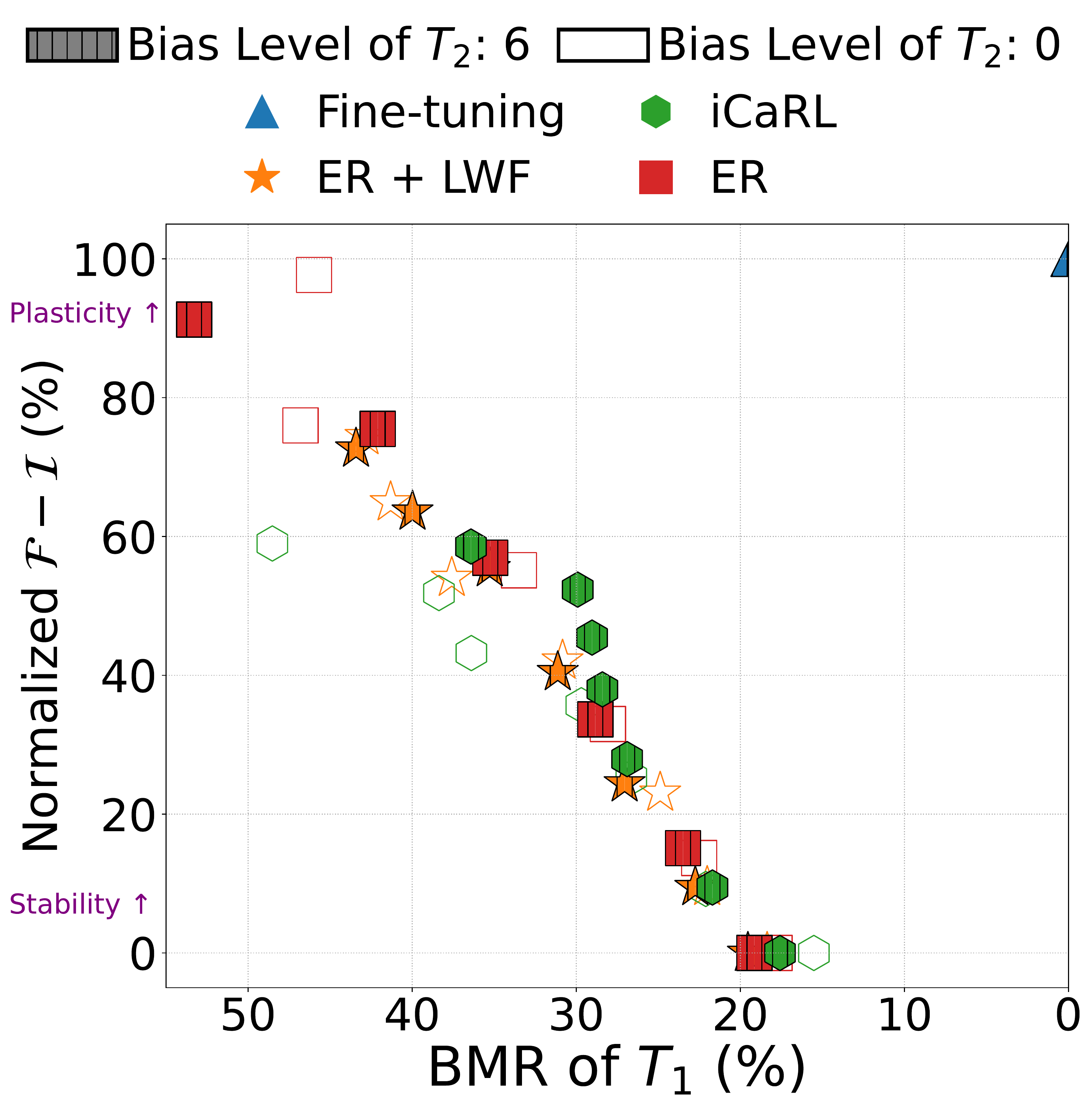}
        \caption{Split ImageNet-100 in Class-IL}
        \label{fig:two_backward_imagenet}
    \end{subfigure}    
    \vspace{-.05in}
    \caption{\small {\bf Backward transfer of bias in two tasks-continual learning.}}
    \label{fig:two_backward}
    \vspace{-.1in}
\end{figure*}

\subsection{Continual learning and debiasing baselines}
We compare two naive methods and six representative CL methods: \textit{fine-tuning} without any consideration of CL, \textit{model-freezing} with freezing model parameters updated from previous tasks, LWF \cite{li2017learning} and EWC \cite{ewc} for regularization based methods, ER \cite{er} and iCaRL \cite{rebuffi2017icarl} for rehearsal based methods and PackNet \cite{mallya2018packnet} for parameter isolation based methods. We note that each CL method can control the stability-plasticity trade-off by adjusting their own hyperparameters such as the regularization strength, the size of the exemplar memory  or the pruning ratio. We further note that PackNet is designed only for task-IL settings, LWF for task-IL and class-IL, and iCaRL for class-IL settings.
Additionally, we employ a widely used debiasing technique, Group DRO \cite{groupdro}.
For implementation details, please refer to the Appendix.

\subsection{Metrics} 
We utilize \textit{average accuracy} over learned tasks as a metric for CL performance and \textit{Normalized $\mathcal{F} - \mathcal{I}$} as a metric for the relative weight on plasticity and stability. 
The concrete definition of Normalized $\mathcal{F} - \mathcal{I}$ is given below. 
Let $h_t$ and $ h^*_t$ be the classifiers learned up to $T_t$ tasks which are trained by a CL method and the fine-tuning, respectively. 
The forgetting and intransigence measures \cite{Chaudhry2018ECCV,cha2021cpr}, $\mathcal{F}_t$ and $\mathcal{I}_t$, after learning up to task $T_t$ are then defined as follows:
\begin{align}
\mathcal{F}_t &: \frac{1}{t-1} \sum_{j=1}^{t-1} \max_{l\in[t-1]} \operatorname{A}(h_l, \mathcal{D}_j)-\operatorname{A}(h_t, \mathcal{D}_j) \label{eq:F} \\
\mathcal{I}_t &: \frac{1}{t} \sum_{j=1}^{t} \operatorname{A}(h^*_j, \mathcal{D}_j)-\operatorname{A}(h_j, \mathcal{D}_j), \label{eq:I}
\end{align}
in which $\operatorname{A}(h, \mathcal{D}_t)$ is the test accuracy of a model for a test dataset of $T_t$, $\mathcal{D}_t$. Note that the two measures evaluate the stability and plasticity of a CL method, respectively. 
Then, for each CL scenario and method, the differences between the two measures are normalized by the maximum and minimum values of $\mathcal{F}-\mathcal{I}$, which are obtained by varying the hyperparameter of each CL method. Especially, in the case of regularization based methods, the maximum and minimum values of $\mathcal{F}-\mathcal{I}$ mostly correspond to $\mathcal{F}-\mathcal{I}$ of the fixed model and the fine-tuning. Notably, the Normalized $\mathcal{F}-\mathcal{I}$ indicates the model focuses more on stability as the value becomes lower and on plasticity as it becomes higher.

The degree of bias of the model can be evaluated by observing its behavior for predicting a sample when a bias feature of the sample is changed. Formally, we measure a model bias using the bias-flipped mis-classification rate (BMR):
\begin{align}
    \text{BMR} = \frac{\sum_{\{x_i \in \mathcal{D}|h(x_i)=y_i\}} \mathbb{I}(h(x_i^*)\neq y_i)}{|\{x_i \in \mathcal{D}|h(x_i)=y_i\}|},
\end{align}
in which $x_i^*$ is a bias-flipped sample with other features fixed, \eg, changes of presence only for color (for Split CIFAR-100S) or watermark (for Split-ImageNet). Namely, BMR considers the number of falsely predicted samples after only flipping bias features for all correctly predicted samples. 

In most real-world datasets such as CelebA, it might be challenging to generate bias-flipped samples such as transforming a woman image to look like a man. Thus, for CelebA, we use the difference of classwise accuracy (DCA) \cite{berk2021fairness} as a surrogate metric: 
\begin{align}
\small
\text{DCA}(h, \mathcal{D}_{t}) &= \frac{1}{|\mathcal{Y}_{t}|}\sum_{y\in\mathcal{Y}_{t}}{\max_{a, a'\in\mathcal{A}}} \lvert \operatorname{A}(h, \mathcal{D}_{t}^{y,a}) - \operatorname{A}(h, \mathcal{D}_{t}^{y,a'}) \rvert,  \nonumber 
\end{align}
in which $\mathcal{D}^{y,a}_{t}$ is the subset of $\mathcal{D}_{t}$ that is confined to the
samples with class-group label pair $(y,a)$. DCA means the average (over class) of per-class maximum accuracy difference between domains. Informally, DCA is regarded as an approximation of BMR by calculating the difference of predictions in group levels, not in sample levels.

We note that we use BMR for Split CIFAR-100S and ImageNet-100 and DCA for CelebA. Further note that high BMR and DCA correspond to $h$ possessing large bias.

\section{Case for CL with two tasks}
\label{sec:two_task_studies}
We begin our analysis by examining CL scenarios consisting of two tasks in sequence. Our goal is to identify \textit{forward} and \textit{backward} transfers of the bias of a CL model through both quantitative analyses and figure out how the CL methods promote each of these transfers. 

\subsection{Forward transfer of bias}
To investigate the forward transfer of the bias, we evaluated CL methods by varying the degree of bias of $T_1$, while that of $T_2$ is fixed to level 0. 
Figure \ref{fig:two_forward} reports bias metric values of $T_2$ along with Normalized $\mathcal{F}-\mathcal{I}$ on three datasets after learning $T_2$ with two different bias levels of $T_1$, \ie, level 0 \& 6. 
In the figure, we plot the results of each CL method by varying their own hyperparameter for controlling the stability-plasticity trade-off. The upper point on each plot represents a lower regularization strength, a smaller memory size, or a lower pruning ratio. 

The following are our observations from the figures.
First, from the gap of blue triangles in Split CIFAR-100S and Split ImageNet-100 results, we observe that even with simple fine-tuning, the bias of $T_1$ adversely affects one of $T_2$, \ie, forward transfer of bias exists, which is consistent with Salman \etal \cite{salman2022does}. However, the overlapping blue triangles on CelebA show that the bias of $T_1$ does not always persist when not considering the stability, implying that the degree of forgetting for the bias of a model acquired from previous tasks can be different depending on the types of bias.
Second, we observe that when applying CL methods, the gap between colored and uncolored points for similar Normalized $\mathcal{F}-\mathcal{I}$ is mostly larger than fine-tuning. Moreover, the gap increases as Normalized $\mathcal{F}-\mathcal{I}$ becomes lower. Namely, these results clearly demonstrate that CL methods promote the forward transfer of bias because they tend to remember the knowledge of past tasks even if it contains some bias features. Furthermore, the extent of the transfer is increasing as CL methods focus on stability more. 
Finally, we observe that bias of $T_2$ is mostly better when learned after $T_1$ with bias level 0 than with bias level 6, under similar Normalized $\mathcal{F}-\mathcal{I}$. Therefore, we argue that before learning a new task, biases of a CL model obtained from previous tasks should be mitigated for learning the new tasks correctly. 

We additionally report the results on Split CIFAR-100S when the bias level of $T_2$ is 2 or 4 in Appendix, and observe similar trends from Figure \ref{fig:two_forward}.


\subsection{Backward transfer of bias}
Now, we investigate the backward transfer of bias. Figure \ref{fig:two_backward} compares the bias of a model at $T_1$ by varying the bias of $T_2$, while the bias level of $T_1$ is fixed as level 0. We omit the results for PackNet, as it freezes the parameters updated in the previous tasks and thereby the predictions for previous tasks are not changed, \ie, any backward transfer does not occur. 

Figure \ref{fig:two_backward_cifar} and \ref{fig:two_backward_celeba} show the opposite trend of our previous results. Firstly, we observe that the bias gap for $T_1$ under similar Normalized $\mathcal{F}-\mathcal{I}$ is maximized by fine-tuning and minimized by model-freezing. Also, for each CL method, the gap becomes severer as the Normalized $\mathcal{F}-\mathcal{I}$ increases. This means that the more CL methods prioritize plasticity over stability, the more bias obtained from $T_2$ is transferred to $T_1$, i.e., the backward transfer of bias occurs more. Additionally, when the bias level of $T_2$ is 0, $T_1$ consistently exhibits lower bias, highlighting the need to address the potential bias of new incoming tasks. 

We identify that BMRs between colored and uncolored points are nearly identical in Figure \ref{fig:two_backward_imagenet}. To reason about this phenomenon, we analyzed the predictions from models and found out that it may be due to the old-new bias which is an inherent issue of Class-IL --- namely, predictions are biased towards new classes due to the imbalance between old and new task data samples. In other words, although $T_2$ does not contain the carton class, the converted image containing watermarks can be predicted to one of the new classes, leading to no difference in BMR. Please refer to Appendix for the related experimental results and more discussions. 

\noindent\textbf{Remarks on developing a new CL method}. 
From two-task analyses in Figure \ref{fig:two_forward} and \ref{fig:two_backward}, we observed that the bias of each task negatively affects the other tasks even if they do not contain the bias. This appeals that whenever we encounter a bias of an incoming task, we should consider learning the task without the bias and preventing forgetting of the previous tasks at the same time. As a straightforward solution, one may naively consider applying an existing debiasing technique (\eg, Group DRO) to the model obtained after learning a current task by a CL method. However, we can easily expect that the accuracy of previous tasks significantly drops whereas the bias of the current task can be successfully reduced
(we report the results of this scenario in Appendix). Hence, we argue that it is necessary to develop a novel approach for taking debiased learning into account while continual learning.

 \begin{figure}
    \centering
    \includegraphics[width=0.6\linewidth]{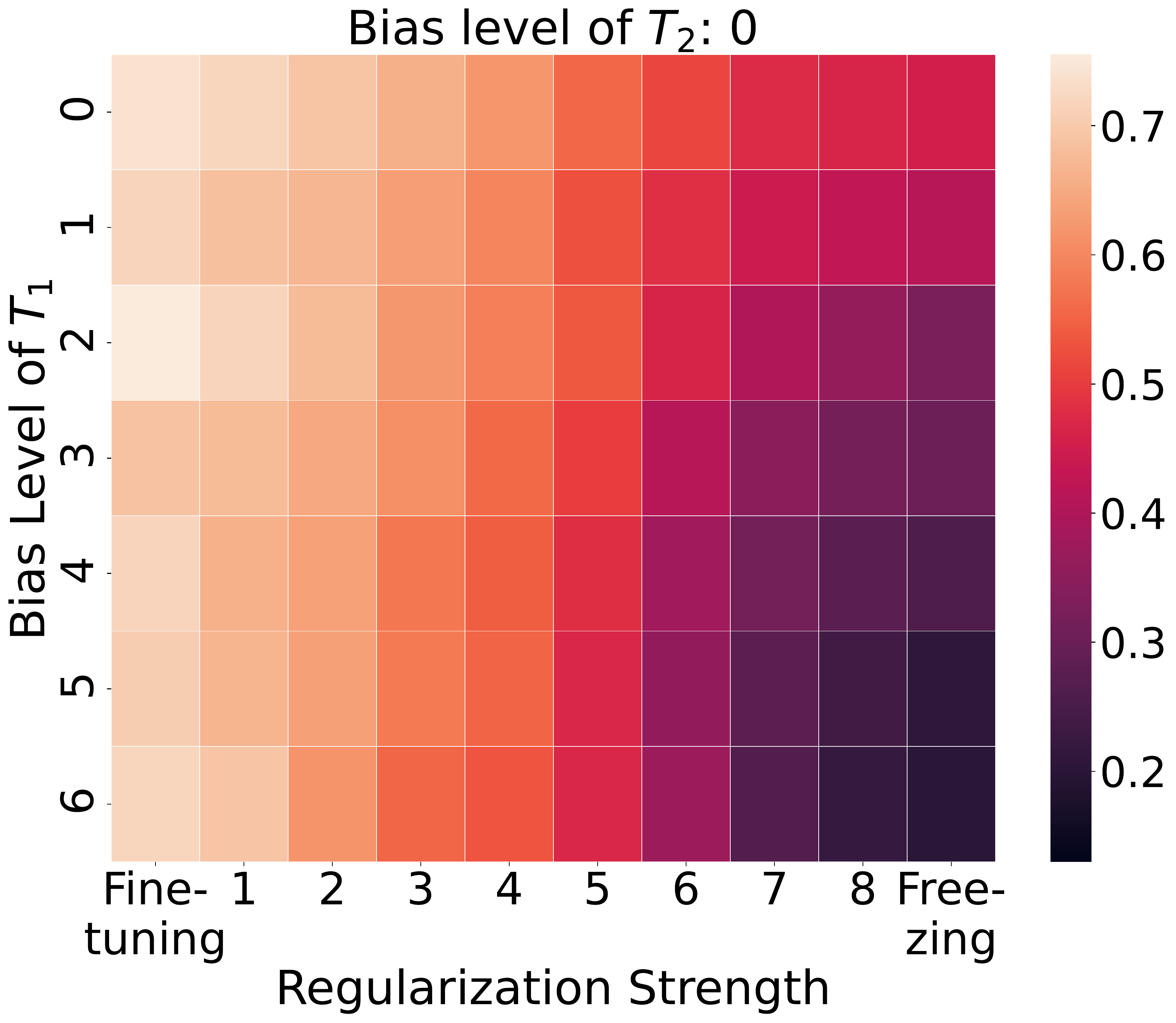}
    \caption{\small\textbf{CKA on Split CIFAR-100S.} To observe the forward transfer of bias, the CKA between color and grayscale images in $T_2$ is shown according to the bias level of $T_1$ and the regularization strength, after learning up to $T_2$ by EWC.}
    \label{fig:cka_forward}
    \vspace{-.1in}
\end{figure}

\subsection{Feature representation analysis}
To provide more direct evidence of bias transfer, we analyze feature representations extracted from the penultimate layer of a DNN-based model using centered kernel alignment (CKA) with the linear kernel \cite{kornblith2019similarity}. CKA is an isotropic scaling-invariant metric for measuring the similarity between two representations of a model. The two plots in Figure \ref{fig:cka_forward} compare the CKA values on Split CIFAR-100S for EWC under the two-tasks settings similar to Figure \ref{fig:two_forward_cifar}. Namely, we evaluate models after learning $T_2$ by varying the regularization strength and the bias level of $T_1$. We then compute the CKA similarity between color and grayscale images in the test dataset of $T_2$. That means, a low CKA value indicates that representations for each group are different, \ie, the model possesses the color bias more. Figure \ref{fig:cka_forward} clearly shows that as the regularization strength is stronger and the bias of $T_1$ is more severe, CKA values decrease. Thus, we also observe the forward transfer of the bias through the analysis of feature representations. The CKA result showing the backward transfer is reported in Appendix.

\section{Case for CL with a longer sequence of tasks}
\label{sec:longer}
\begin{figure}[t!]
    \centering
    \begin{subfigure}[t]{\linewidth}
        \centering
        \includegraphics[width=0.7\linewidth]{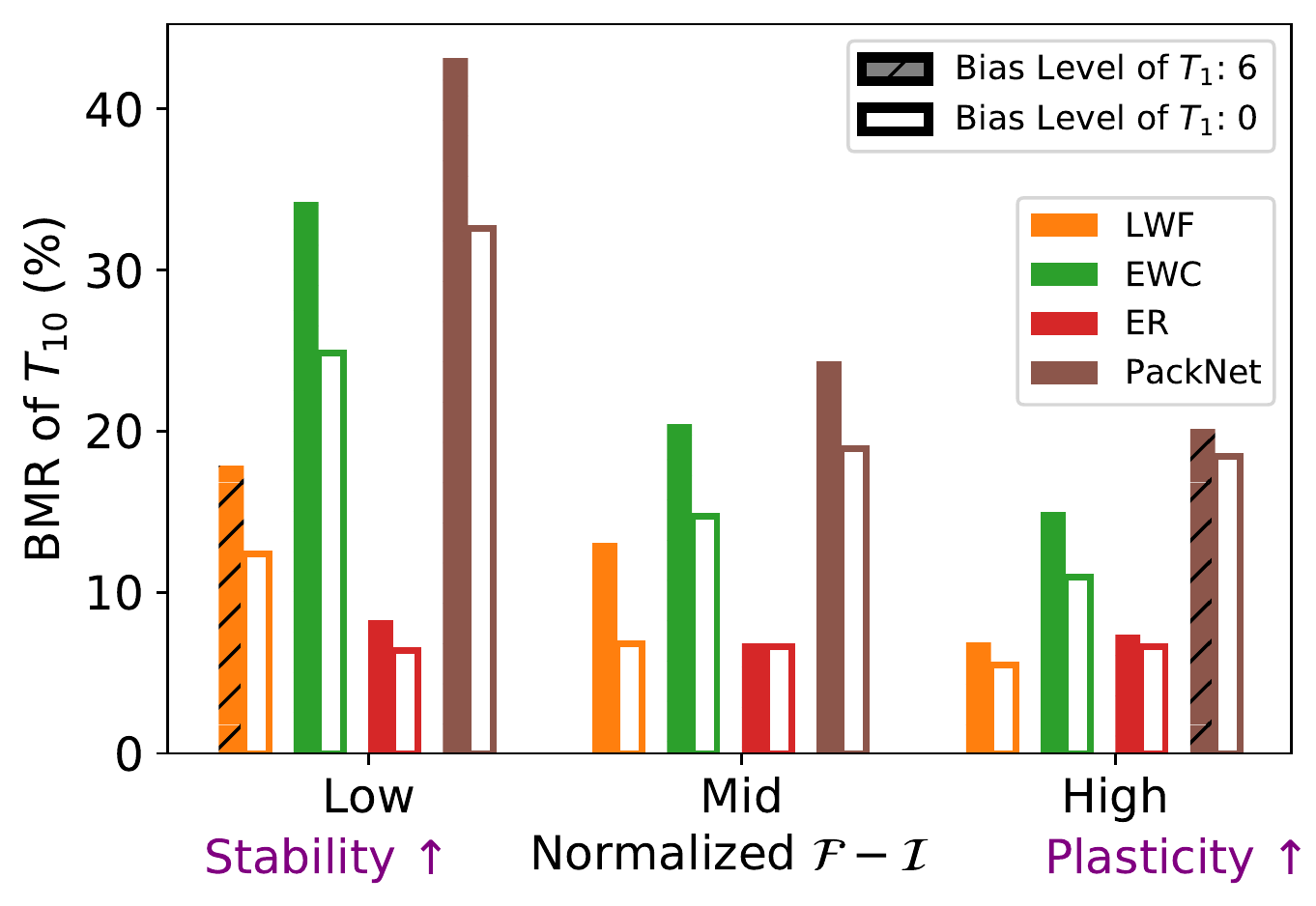}
        \caption{\small {Forward transfer of bias}}
        \label{fig:long_forward_cifar}
    \end{subfigure}
    \begin{subfigure}[b]{\linewidth}
        \centering
        \includegraphics[width=0.7\linewidth]{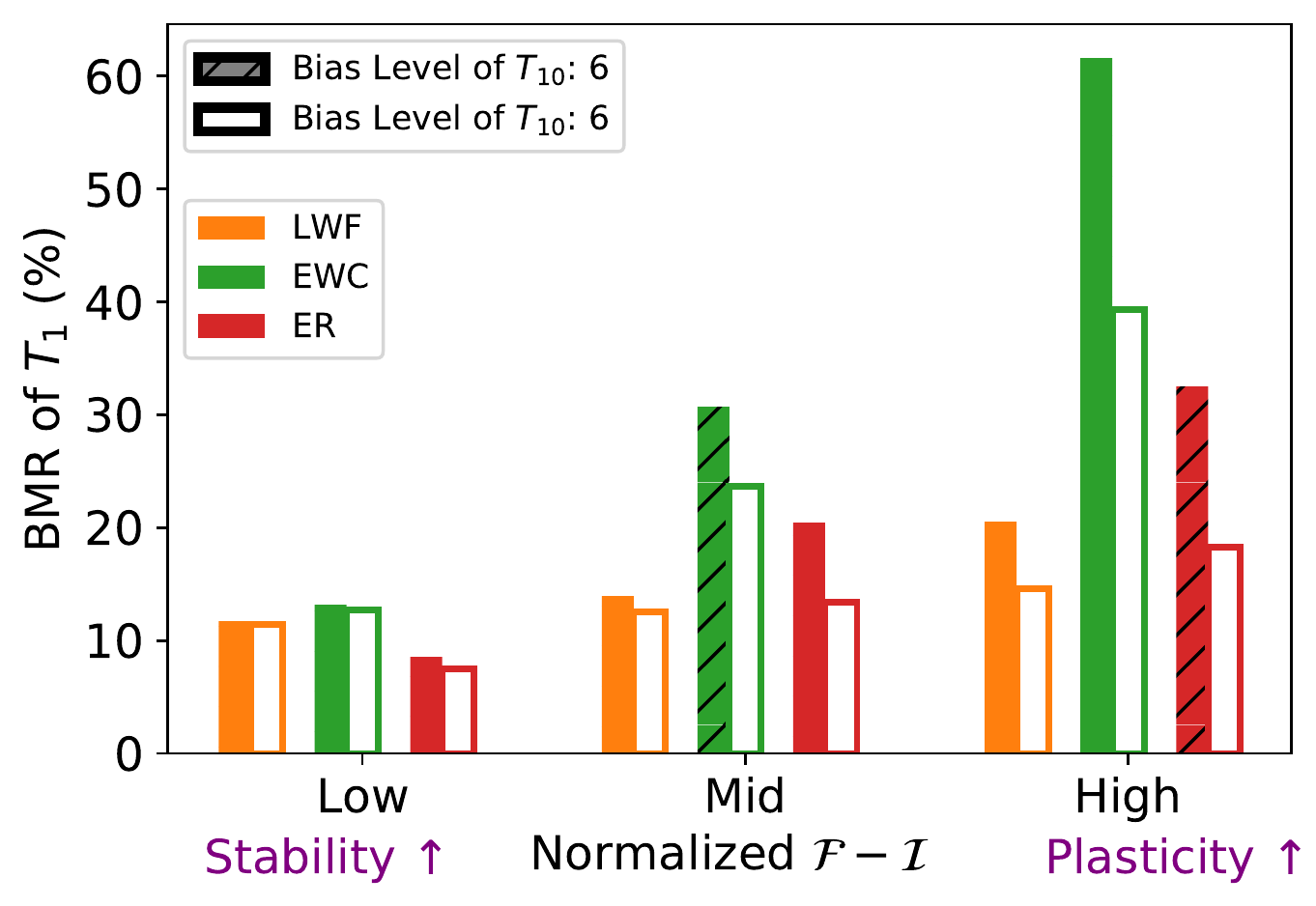}
        \caption{\small {Backward transfer of bias}}
        \label{fig:long_backward_cifar}
    \end{subfigure}
    \caption{\small {\bf Bias transfers in a sequence of 10 tasks on Split-CIFAR100S}. The BMRs of $T_{10}$ or $T_1$ are shown after learning up to $T_{10}$ by CL methods.}
    \label{fig:long_cifar}
    \vspace{-.1in}
\end{figure}

We confirmed in the previous section that the bias is transferred forward and backward in tow-tasks CL scenarios. In this section, we further investigate the bias transferability in a sequence of multiple tasks. 

We note that for all figures in this section, we simplify the visual format for better clarity of the comparison; in detail, we divide a range of the normalized $\mathcal{F}-\mathcal{I}$ into three equal intervals and report a result for each interval. Given a CL scenario, we evaluate CL methods several times for varying their hyperparameters and select one of the results with the highest average accuracy for each interval. 

\subsection{Persistence of bias transfer in longer sequences}
\label{subsec:long_transfer}
Firstly, we consider a sequence of 10 tasks on Split CIFAR-100S to observe that the bias transfer exists in longer CL scenarios.
Similar to settings in Figure \ref{fig:two_forward}, we vary the bias level of the first or last task with level 0 or 6, while the bias level of all other tasks is fixed as level 0. The two plots in Figure \ref{fig:long_cifar} show BMR of $T_{10}$ (resp. $T_1$) according to the bias of $T_1$ (resp. $T_{10}$) for each CL method. 

Figure \ref{fig:long_cifar} reveals an analogous trend with two-task analyses. Specifically, the two plots exhibit the BMR of $T_{10}$ (resp. $T_1$) always higher when the bias level of $T_1$ (resp. $T_{10}$) is 6.
Furthermore, the gap of BMR between them is at its widest when the normalized $\mathcal{F}-\mathcal{I}$ is low (resp. high). 
Namely, the bias transfers also occur in longer sequences of tasks. 
We additionally display the accuracy of $T_{10}$ and $T_{1}$ for each plot in Appendix and find that the accuracies are roughly the same, meaning that the gap of BMR is due to the bias transfer. 
Finally, we emphasize that from Figure \ref{fig:long_forward_cifar}, the color bias of $T_1$ can \textit{persist} even if a CL model learns nine additional tasks with the bias level of 0, especially when focusing on the stability.

\subsection{Accumulation of the same type of bias}
We now verify whether the \textit{same} type of bias of each task would be \textit{accumulated} by CL methods; namely, when the number of previous tasks with the same type of dataset bias is increasing, the CL models make much more biased predictions for the current task. To this end, we take a sequence of 5 tasks with a bias level of 0 in Split CIFAR-100S. We then randomly select some of the four tasks except the last task and change the bias level of them to 4.

\begin{figure}
    \centering
    \includegraphics[width=0.75\linewidth]{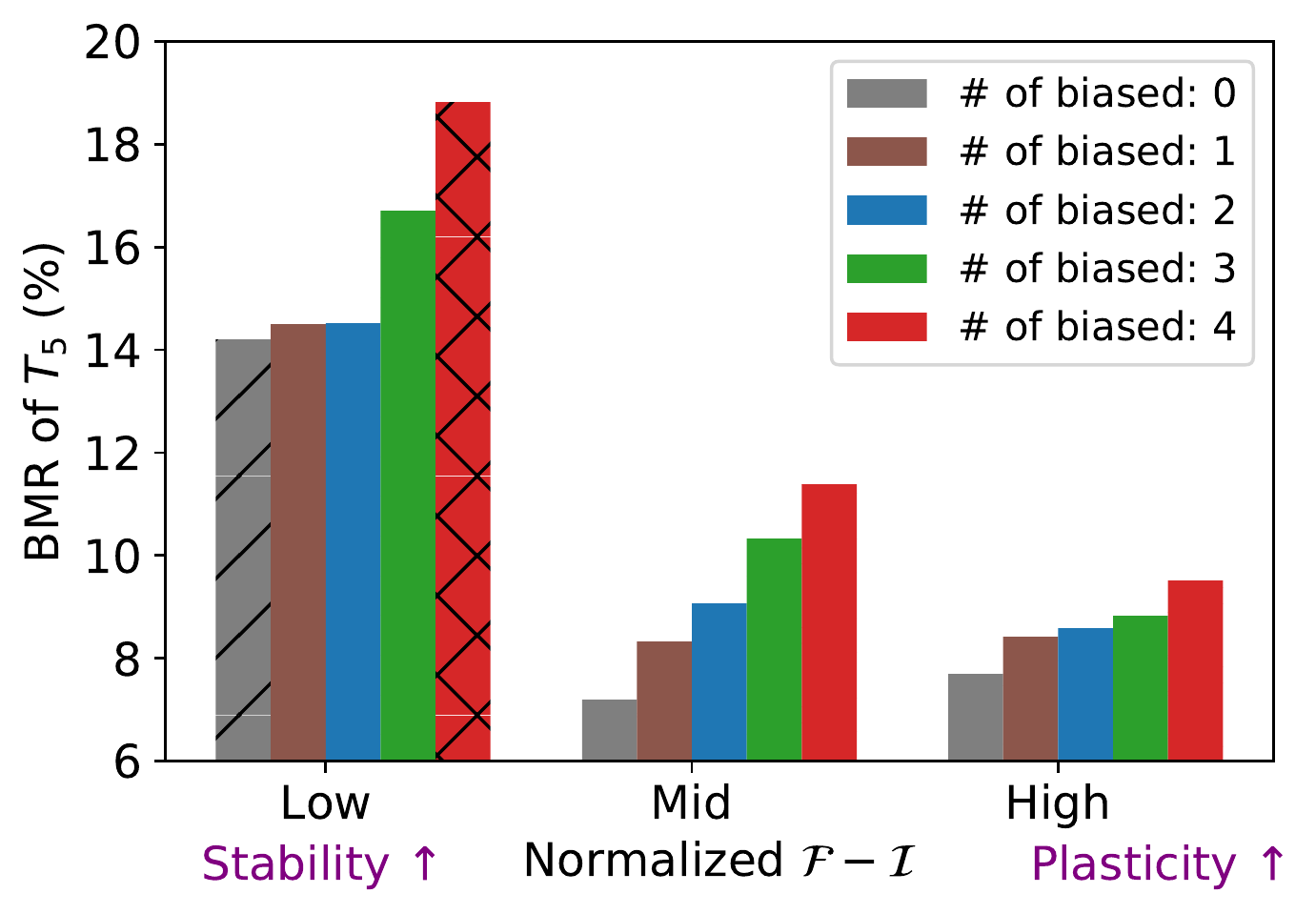}
    \caption{\small {\bf Accumulation of the same type of bias on Split CIFAR-100S.} BMR of $T_5$ is shown depending on the number of biased tasks after up to learning $T_5$ by LWF.}
    \label{fig:accumul_cifar}
    \vspace{-.2in}
\end{figure}

Figure \ref{fig:accumul_cifar} reports BMR values for LWF depending on the number of biased tasks. We clearly check that the BMR of $T_5$ increases as the number of biased tasks increases, and the increase is more significant in the low and middle ranges. Namely, it demonstrates that when tasks with the same type of bias are continuously upcoming, their biases accumulate in the CL model. In addition, we observe that when the number of biased tasks is low in the low range, the gap of BMR is small. It would be because if only the tasks in the middle of sequences have the dataset bias, their dataset bias could not be sufficiently learned under the low plasticity.

\subsection{Accumulation of the different types of bias}
In order to investigate that the \textit{different} type of biases can also accumulate, we consider sequences of three tasks, each randomly picked from CelebA$^8$. We suppose that each task can include one of two kinds of dataset bias. That is, in the training datasets of each task, the class attribute, ``young'', can spuriously correlate with one of two group attributes, ``gender'' or `smiling''. Then, when bias levels of $T_3$  for both group attributes are fixed to 0, we compare the BMRs of $T_3$ depending on whether or not $T_1$ and $T_2$ have gender or smiling biases, respectively. 

\begin{figure}
    \centering
    \begin{subfigure}{0.7\linewidth}
        \centering
        \includegraphics[width=\linewidth]{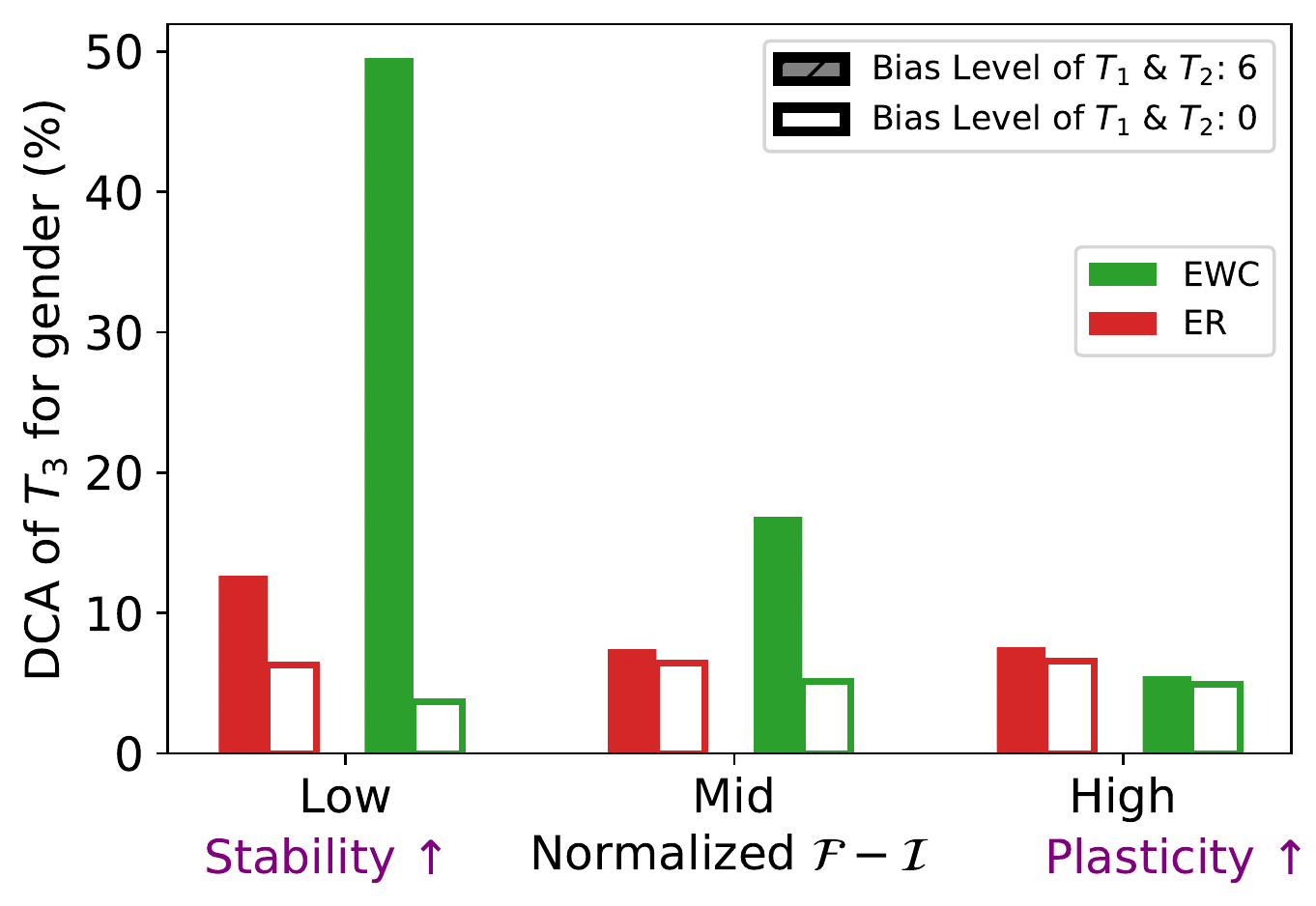}
        \caption{\small{DCA of $T_3$ for the gender bias}}
        \label{fig:accumul gender bias}
    \end{subfigure}
    \begin{subfigure}{0.7\linewidth}
        \centering 
        \includegraphics[width=\linewidth]{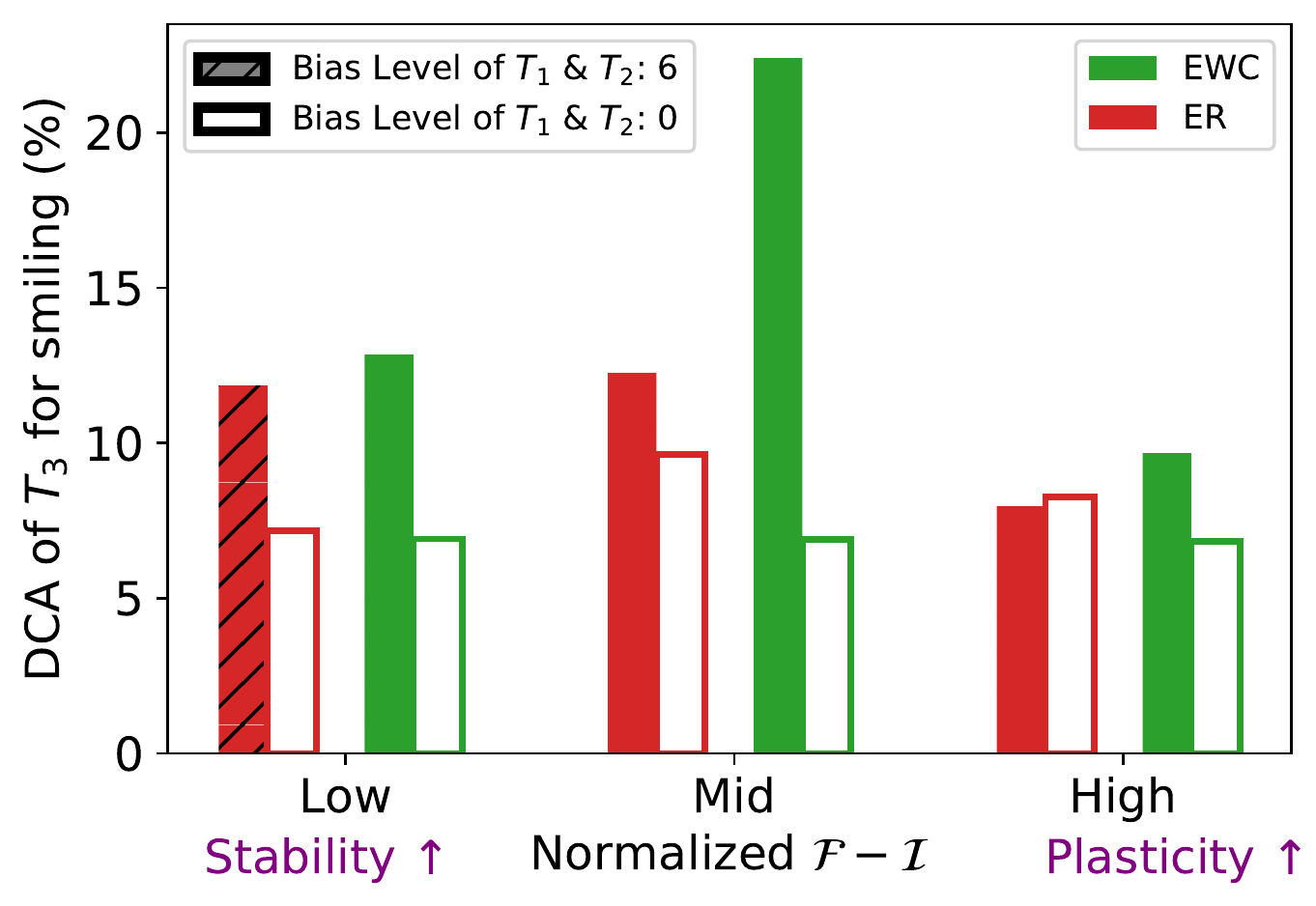}
        \caption{\small{DCA of $T_3$ for the smiling bias}}
        \label{fig:accumul smiling bias}
    \end{subfigure}
    \caption{\small {\bf Accumulation of different types of bias on CelebA$^8$.} DCAs of $T_3$ for ``gender'' and ``smiling'' attributes are reported.}
    \label{fig:accumul_celeba}
    \vspace{-.1in}
\end{figure}

The results are shown in Figure \ref{fig:accumul_celeba}, which displays the degree of gender and smiling bias at $T_3$ for EWC and ER. We find that when $T_1$ and $T_2$ are biased towards gender and smiling, respectively, BMR at $T_3$ are much higher in terms of both group attributes. Moreover, in most cases, we again observe the stronger forward transfer of bias when CL methods focus on the stability more. In Figure \ref{fig:accumul smiling bias}, the gap between colored and uncolored bars for EWC is bigger in the middle range, compared to the low range. We infer that this would be because the dataset bias of $T_2$ could be learned more under higher plasticity, resulting in the bias being more transferred to $T_3$. 

\section{Bias-aware Continual learning} 
\label{sec:comparison}
We demonstrated that the bias of each task can be transferred by naive CL methods in various continual learning situations, highlighting the need for a new approach to continual learning that considers debiasing each task. To address this need, we propose a simple, yet efficient baseline method, dubbed as \methodnamefull (\ours), that can be easily combined with existing CL methods while preventing forgetting of the learned tasks.

\subsection{\methodnamefull (\ours)}
Our proposed method is inspired by a recent CL and debiasing method, specifically GDumb (Greedy Sampler and Dumb Learner) \cite{prabhu2020gdumb} and DFR (Deep Feature Re-weighting) \cite{kirichenko2022last}. GDumb greedily stores a small number of data into an exemplar memory equally for each class in all tasks and uses them to train a model from scratch during the test time. By learning all tasks at the same time, GDumb can improve the average accuracy over learned tasks. On the other hand, DFR retrains only the classification head using a small number of group-balanced data after training a model from scratch using an entire training data that may contain any dataset bias. The authors demonstrate that even if the model learned the biased feature  representations, DFR can remove them by only re-training the last layer using a small portion of group-balanced data. We emphasize that both methods are simple to implement and easily applicable to various settings, and can achieve better or comparable performance for CL or debiasing, respectively, compared to more complex and recent methods in its literature.

\begin{table}[t]
\caption{\small {\bf The comparison of methods on Split CIFAR-100S.} The average accuracy and BMR over 10 tasks are shown. The $k$ denotes the size of exemplar memory, and the numbers in parentheses stand for the standard deviations of each result obtained with different seeds. }
\centering
\resizebox{0.85\columnwidth}{!} {
\begin{tabular}{lcc}
\toprule
Method & Avg. Acc. (\%) & Avg. BMR (\%) \\
\midrule
Fine-tuning     & 19.49 (1.46)                & 51.96 (10.96)          \\
+ BGS (k=1000)   & \textbf{42.75 (2.27)}                & \textbf{34.38 (5.70)}           \\
+ BGS (k=2000)   & \textbf{47.44 (2.80)}                & \textbf{30.42 (5.71)}           \\
\midrule
LWF \cite{li2017learning}             & 59.61 (2.04)                & 28.00 (2.38)           \\
+ BGS (k=1000)   & \textbf{65.00 (1.20)}                & \textbf{18.18 (0.84)}           \\
+ BGS (k=2000)   & \textbf{66.88 (1.02)}                & \textbf{16.82 (0.90)}           \\
\midrule
EWC \cite{ewc}            & 35.57 (1.71)                & 46.65 (2.31)           \\
+ BGS (k=1000)   & \textbf{46.20 (1.08)}                & \textbf{31.65 (2.12)}           \\
+ BGS (k=2000)   & \textbf{49.30 (1.37)}                & \textbf{28.49 (1.84)}           \\
\midrule
ER (k=1000) \cite{er}    & 54.77 (1.68)                & 29.17 (3.81)           \\
+ BGS        & \textbf{59.38 (0.98)}                & \textbf{20.89 (2.28)}           \\
ER (k=2000)     & 60.75 (1.12)                & 25.47 (1.41)           \\
+ BGS    & \textbf{65.00 (0.74)}                & \textbf{17.62 (1.46)}           \\
\midrule
PackNet \cite{mallya2018packnet}            & 47.46 (1.52)                & 33.97 (6.10)           \\
+ BGS (k=1000)   & \textbf{47.69 (2.88)}                & \textbf{31.26 (3.93)}           \\
+ BGS (k=2000)   & \textbf{49.39 (2.86)}                & \textbf{29.15 (4.05)}           \\
\midrule
GDumb (k=1000) \cite{prabhu2020gdumb}  & 32.33 (1.98)                & 52.15 (4.95)           \\
GDumb (k=2000)  & 41.31 (1.99)                & 45.43 (3.56)           \\
\midrule
\midrule
LWF + Group DRO \cite{groupdro} & 59.39 (1.41)                & 23.35 (2.72)          \\
\bottomrule
\end{tabular}
}
\vspace{-.1in}
\label{tab:cifar}
\end{table}

In a similar spirit, the algorithm of our \ours is two steps: first, during CL using a employed typical CL method, \ours stores group-class balanced data over all seen classes and groups in a greedy manner as the same process as GDumb, which shown is in Appendix. \ours then retrains only the classification heads of a neural network trained by the CL method by using the group-class balanced exemplar memory. By doing this process, we expect that \ours can mitigate the bias of the model like DFR while preventing forgetting of previous tasks. Moreover, we emphasize that \ours can be used in conjunction with any existing CL method without any additional hyperparameters. 

\subsection{Performance comparison}
We evaluate our \ours using 10-task sequences on Split CIFAR-100S and 8-task sequences on CelebA$^8$. Each task has a random bias level ranging from 0 to 6 (we did not conduct experiments on Split ImageNet-100 due to the lack of group labels (\ie, watermark labels) in the training dataset). We compare the standard CL methods including GDumb, and combinations with the CL methods and \ours. Additionally, we evaluate naive combination with the best performing regularization CL method for each dataset and a debiasing technique, Group DRO. We tuned the hyperparameters for each CL method and Group DRO based on the average accuracy and BMR up to $T_3$, following the hyperparameter selection protocol used in \cite{mai2022online} and used them to learn the rest of the tasks. 

Table \ref{tab:cifar} and \ref{tab:celeba} present the average accuracy and BMR over all tasks for each method on Split CIFAR-100S and CelebA$^8$. We evaluate replay based methods, ER, GDumb, and \ours, with the two kinds of memory size, which correspond to 10 or 20 images per class, respectively. From the tables, we first observe that applying \ours into the standard CL methods leads to improve BMR in all cases and the average accuracy on Split CIFAR-100S, while slightly dropping the average accuracy on CelebA. In addition, we obviously see that the performance gain by \ours in terms of CL and debiasing performances increases when using a large exemplar memory. We emphasize that such improvements from \ours require any additional hyperparameter tuning for debiasing. On the other hand, although applying Group DRO to CL methods shows good performance on CelebA$^8$ in terms of BMR, it needs additional hyperparameter tuning for Group DRO, which may be prohibitive for practice. Moreover, we observet that its improvement is marginal on Split CIFAR-100S since LWF + Group DRO  may fail to find a good hyperparameter for Group DRO when tuning it with only three tasks, not ten tasks. 

\begin{table}[t]
\caption{\small {\bf The comparison of methods on CelebA$^8$.} The other settings are identical to Table \ref{tab:cifar}. If there is an improvement by \ours, the result is shown in bold.}
\centering
\resizebox{0.9\columnwidth}{!} {
\begin{tabular}{lcc}
\toprule
Method          & Avg. Acc. (\%) & Avg. DCA (\%) \\
\midrule
Fine-tuning     & 79.77 (1.06)        & 34.31 (7.53)  \\
+ BGS (k=320)    & 79.62 (1.24)       & \textbf{31.42 (7.04)}   \\
+ BGS (k=640)    & 79.33 (1.86)        & \textbf{30.30 (6.95)}   \\
\midrule
EWC \cite{ewc}         & 80.19 (1.4)        & 36.30 (9.8)   \\
+ BGS(k=320)     & 79.15 (2.10)        & \textbf{33.87 (11.74)}   \\
+ BGS (k=640)    & 78.98 (1.72)        & \textbf{32.70 (11.64)}   \\
\midrule
ER (k=320) \cite{er}     & 80.99 (0.74)        & 37.68 (6.51)   \\
+ BGS (k=320)    & 80.19 (1.89)        & \textbf{31.04 (5.11)}   \\
ER (k=640)      & 81.00 (0.80)        & 37.50 (5.92)   \\
+ BGS (k=640)    & 79.98 (1.50)        & \textbf{30.49 (2.54)}   \\
\midrule
GDumb (k=320) \cite{prabhu2020gdumb}  & 69.38 (1.19)        & 36.45 (5.54)   \\
GDumb (k=640)   & 72.55 (1.26)        & 42.25 (1.48)   \\
\midrule
\midrule
EWC + Group DRO \cite{groupdro} & 75.32 (3.58)        & 21.79 (3.98)  \\
\bottomrule
\end{tabular}
}
\label{tab:celeba}
\vspace{-.1in}
\end{table}

\noindent \textbf{Remarks for limitations}. 
Although we demonstrated that \ours can improve average bias metric values in CL scenarios without requiring additional hyperparameter, it is important to note that our method does not fundamentally solve the bias-aware CL problem. Namely, even with \ours, the feature representation of a CL model could be still biased. Furthermore, \ours does not work in the absence of group labels in the training dataset, such as Split ImageNet-100. Nevertheless, we hope that \ours serves the purpose of a standard baseline for the bias-aware CL problems.

\section{Concluding remark}
With systematical analyses for two-task and multiple-task CL scenarios, we showed the bias can be transferred both forward and backward by typical CL methods that are oblivious to the dataset bias. 
Furthermore, we appealed to the CL research community to pay attention to the bias-aware CL problem by exhibiting our in-depth analyses, and proposed a simple baseline method for this problem. For future work, we will develop a more principled approach which can accomplish continual learning and debiasing simultaneously. 
{\small
\bibliographystyle{ieee_fullname}
\bibliography{references}
}

\clearpage
\onecolumn

\appendix
\appendix
\numberwithin{equation}{section}
\numberwithin{figure}{section}
\numberwithin{table}{section}
\section*{Appendix}
We offer supplementary materials in this document. Specifically, we provide the detailed algorithm of \ours in Section \ref{sec_append:algo}. In Section \ref{sec_append:details}, we present implementation details including model architectures, optimization, implementations of baseline methods, and the range of hyperparameters used. In addition, we report some additional results for a naive debiasing scenario in Section \ref{sec_append:naive_debiasing} and other various settings in Section \ref{sec_append:additional_results}. Finally, we discuss the results of our study for the backward transfer on Split ImageNet-100 in Section \ref{sec_append:discussion}.
\section{\ours sampling algorithm}
\label{sec_append:algo}
\begin{algorithm}
    \caption{\methodnamefull}
    \label{alg:ours}
    \SetKwInOut{Input}{Input}
    \SetKwInOut{Output}{Output}
    \SetKwInOut{Init}{Init}
    \Init{Memory $\mathcal{M}=\{\{\}, \dots, \}$ with capacity $K$, Labels $\mathcal{L}=\{\}$, Count $C=\{0,\dots\}$ 
    }
    \Input{a data sample $(x_i, a_i, y_i)$, task id $t$}
    
    \Output{$\mathcal{M}$}
    $k \gets \frac{K}{|\mathcal{L}| \times |\mathcal{A}|}$\;
    \If{$C\big[(a_i, y_i, t)\big] < k$}{
    \eIf{$\Sigma_{a\in\mathcal{A}, (y,t)\in\mathcal{L}} C\big[(a,y,t)\big] < K$}{
        $\mathcal{M}\big[(a_i, y_i, t)\big] = \mathcal{M}\big[(a_i, y_i, t)\big] \cup (x_i, a_i, y_i, t)$\;
    } 
    {
        $(a_j, y_j, t_j)$ = $\operatorname{argmax}_{(a, y, t)} C\big[(a,y,t)\big] $\;
        $\mathcal{M}\big[(a_j, y_j, t_j)\big].\operatorname{pop}()$\;
        $\mathcal{M}\big[(a_i, y_i, t)\big] = \mathcal{M}\big[(a_i, y_i, t)\big] \cup (x_i, a_i, y_i, t)$\;
        $C[(a_j, y_j, t_j)] = C[(a_j, y_j, t_j)] - 1$ \;
    }
    \If{$C\big[(a_i, y_i, t)\big]==0$}{
        $\mathcal{L} = \mathcal{L} \cup (y_i, t)$\;
    }
        $C\big[(a_i, y_i, t)\big] = C\big[(a_i, y_i, t)\big] + 1$\;
    }
\end{algorithm}
\section{More implementation details}
\label{sec_append:details}
\subsection{Model architectures and optimization}
For all datasets, we used the AdamW optimizer \cite{adamw} with the following hyperparameters: learning rate of 0.001, weight decay of 0.01, $\beta_1$ of 0.9 , $\beta_2$ of 0.999, and $\epsilon$ of $10^{-8}$. 
For Split CIFAR-100S, we trained ResNet-56 \cite{resnet} from scratch for 70 epochs using a batch size of 256. 
For CelebA and Split ImageNet-100, we trained ResNet-18 from scratch for 50 and 70 epochs, respectively, using a batch size of 128. We incorporated the cosine annealing learning rate decay, with the maximum number of iterations set to the same as the number of training epochs.

\subsection{Implementations of continual learning methods}
In the original Elastic Weight Consolidation (EWC) algorithm \cite{ewc}, the snapshot of a CL model should be stored whenever the model is updated from a new task. This stored model is then used to calculate the importance scores of model parameters in the new task. Namely, the algorithm requires a linearly growing amount of memory to store a sequence of models, which is space-inefficient. To address this issue, we implemented online EWC, proposed in \cite{schwarz2018progress}, which averages the importance scores in an online manner without storing a set of models. 

For Learning Without Forgetting (LWF) \cite{li2017learning}, we used an average of the distillation losses for each head to balance between the cross entropy loss and the distillation losses.

For Experience Replay (ER) \cite{er}, incremental classifier and representation learning (iCaRL) \cite{rebuffi2017icarl} and Packnet \cite{mallya2018packnet}, we implemented the same as their original versions. For ER, we employed the Reservoir sampling \cite{reservoir}  as a strategy for updating the exemplar memory.  

\subsection{Hyperparameters for each result}
In our experiments, we evaluated CL methods several times by varying their hyperparameters based on the sets of candidates. The  candidates are uniformly distributed on a logarithmic scale within a given range. For all figures presented in Section 4 and 5, we then plotted results for some of those hyperparameter candidates. We note that we omitted some overlapped results in Section 4 to enhance visibility. For the table results in Section 6, we tuned the hyperparameters for regularization based methods and Group DRO using the same candidates. We included Table \ref{table:hyperparams} to provide full ranges of hyperparameters tested. 
We note that the memory size specified in Table \ref{table:hyperparams} represents the fraction of the number of training images included in one of all tasks except the last one. 
\begin{table}[t]
    \caption{\small \textbf{Hyperparameter search ranges.}}
    \label{table:hyperparams}
    \centering
    \small
    \begin{tabular}{ccc}
    \toprule
        Method & Hyperparameter & Search range \\ \midrule
        EWC \cite{ewc} & Regularization strength $\lambda$ &[$10^0, 10^9$] \\\midrule
        LWF \cite{li2017learning} & Regularization strength $\lambda$ &[$10^{-2}, 3\times10^{2}$] \\\midrule
        ER \cite{er} & Memory size & [$10^{-3}, 10^{0}$] \\ \midrule
        iCaRL \cite{rebuffi2017icarl} & Memory size & [$10^{-3}, 10^{0}$]  \\ \midrule
        PackNet \cite{mallya2018packnet}  & Pruning ratio $r$ & [$10^{-1}, 8\times10^{-1}$] \\ \midrule
        Group DRO \cite{groupdro}  & Learning rate of $q$ &  [$10^{-8}, 10^2$] \\ 
        \bottomrule
    \end{tabular}
\end{table}
\subsection{Datasets}
\begin{figure*}[t!]
    \centering
    \begin{subfigure}[t]{0.4\linewidth}
        \includegraphics[width=0.9\linewidth]{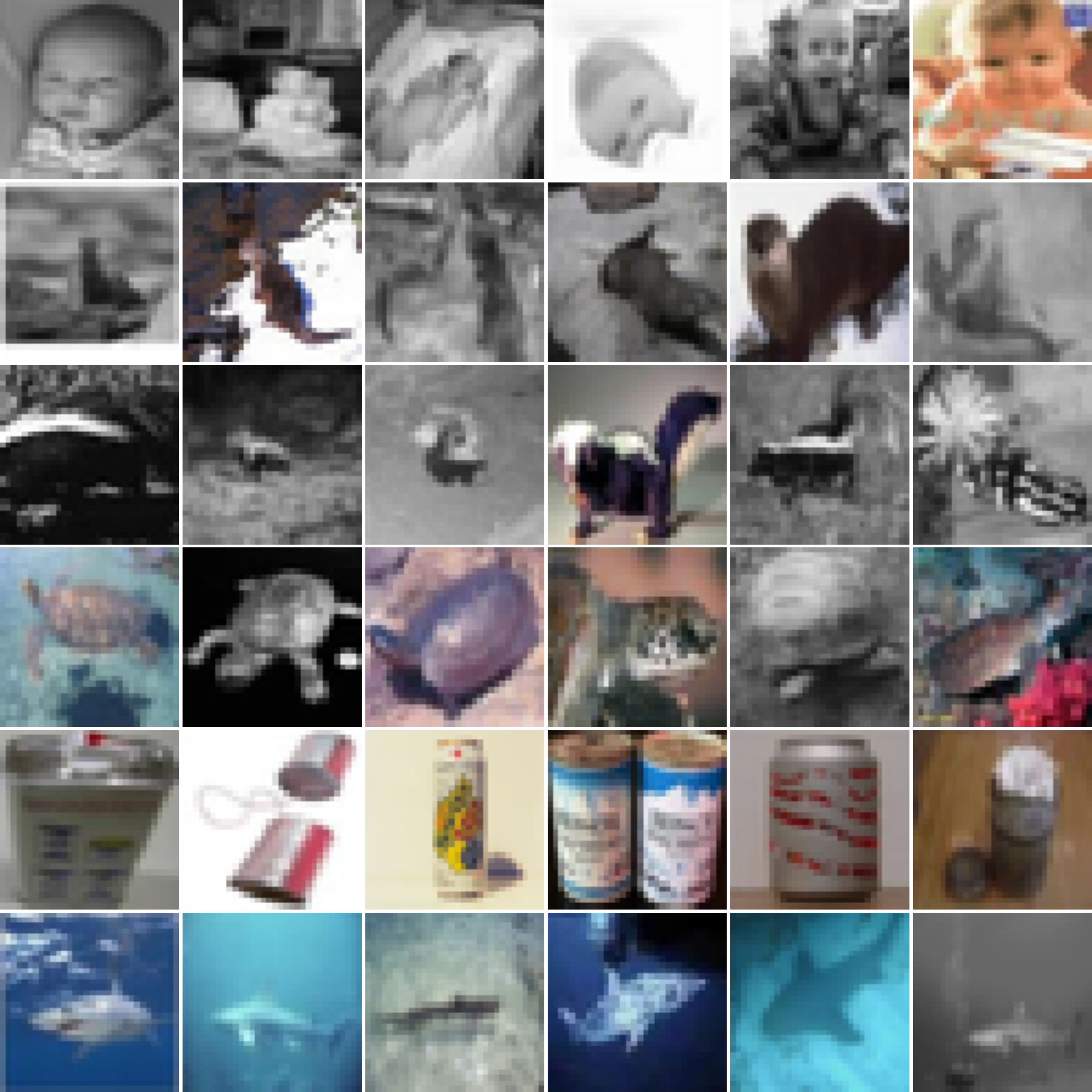}
        \caption{Training samples}
        \label{fig:split_cifar100s_train}
    \end{subfigure}
    \begin{subfigure}[t]{0.4\linewidth}
        \includegraphics[width=0.9\linewidth]{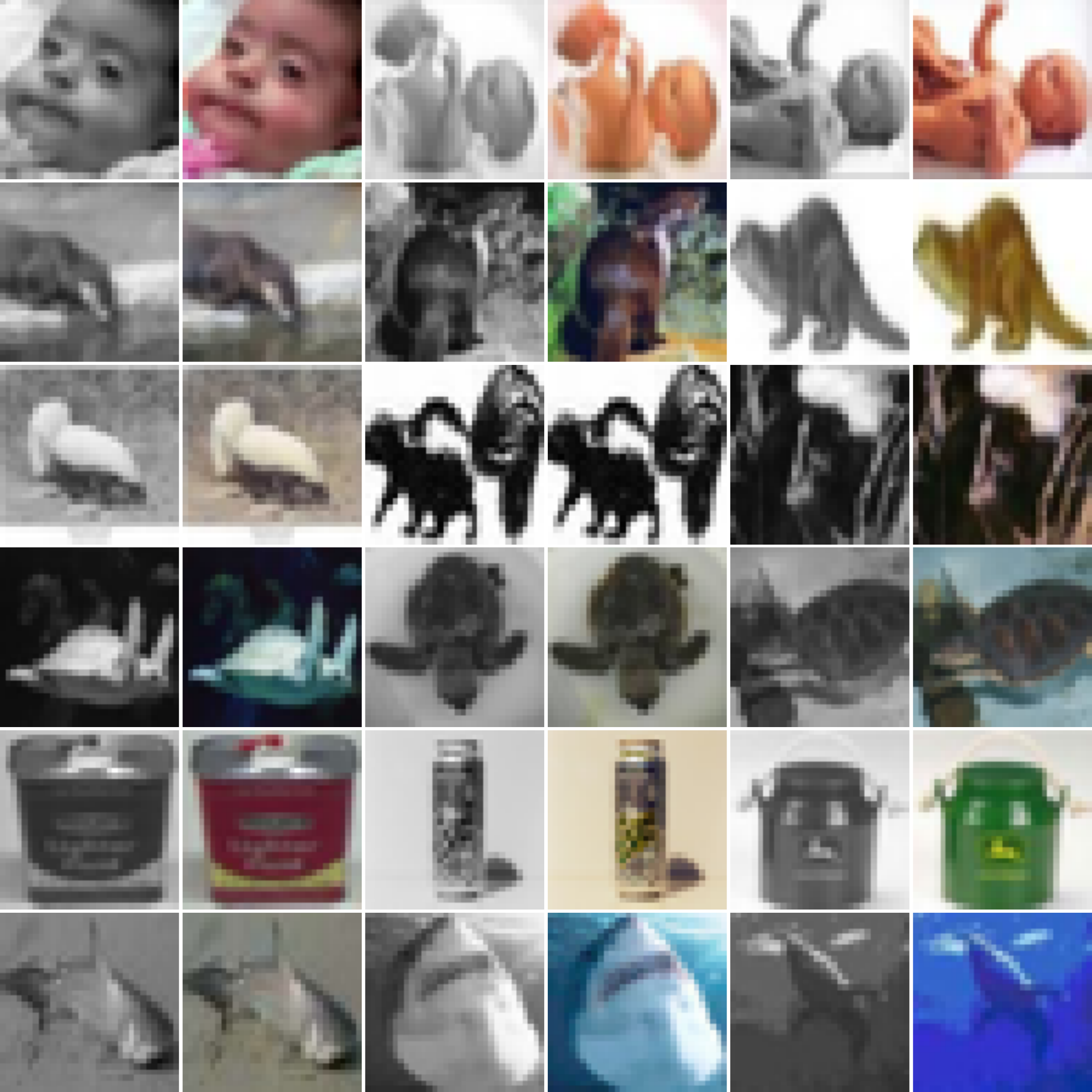}
        \caption{Test samples}
        \label{fig:split_cifar_100s_test}
    \end{subfigure}
    \caption{\small {\bf Samples in a certain task with bias level of 2 in Split CIFAR-100S.} Each row represents a specific class within the task. The top three rows represent classes biased toward the grayscale samples, while the bottom three rows contain classes biased toward the color samples. The test dataset includes pairs of images, where each pair contains one grayscale and one color version of the same image.}
    \label{fig:split_cifar100s_samples}
\end{figure*}

\noindent\textbf{Split CIFAR-100S}. In the training dataset of each task, the first five classes are skewed toward the color group and the latter skewed toward the grayscale group, given the skew ratio. For the test dataset of each task, we have pairs of the same images; ones in color, and ones in grayscale. Fig \ref{fig:split_cifar100s_samples} illustrates some training and test samples in Split CIFAR-100S.

\begin{figure*}[t!]
    \centering
    \begin{subfigure}[t]{0.4\linewidth}
        \includegraphics[width=0.9\linewidth]{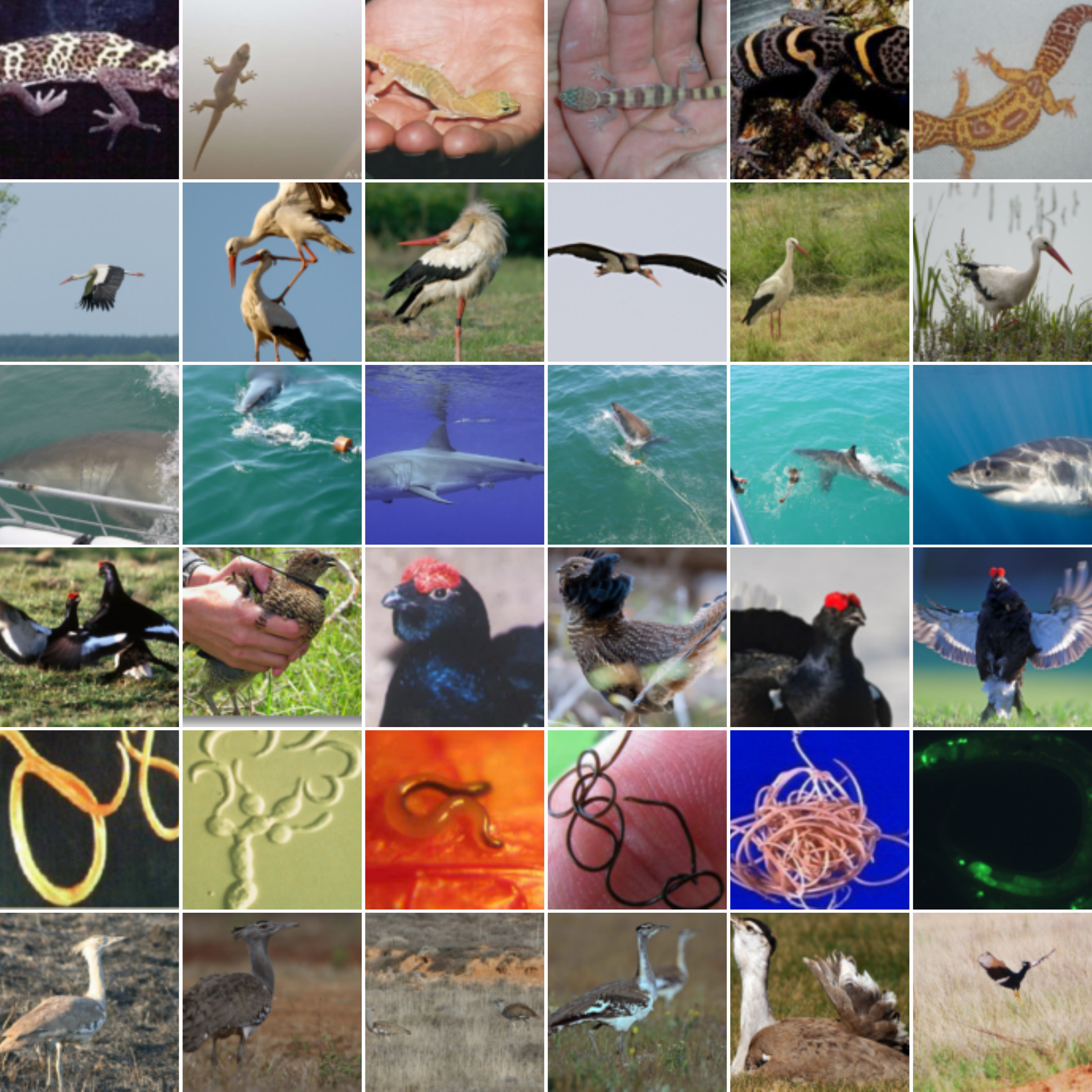}
        \caption{Without ``Carton'' class}
        \label{fig:split_imagenet_train_wo_carton}
    \end{subfigure}
    \begin{subfigure}[t]{0.4\linewidth}
        \includegraphics[width=0.9\linewidth]{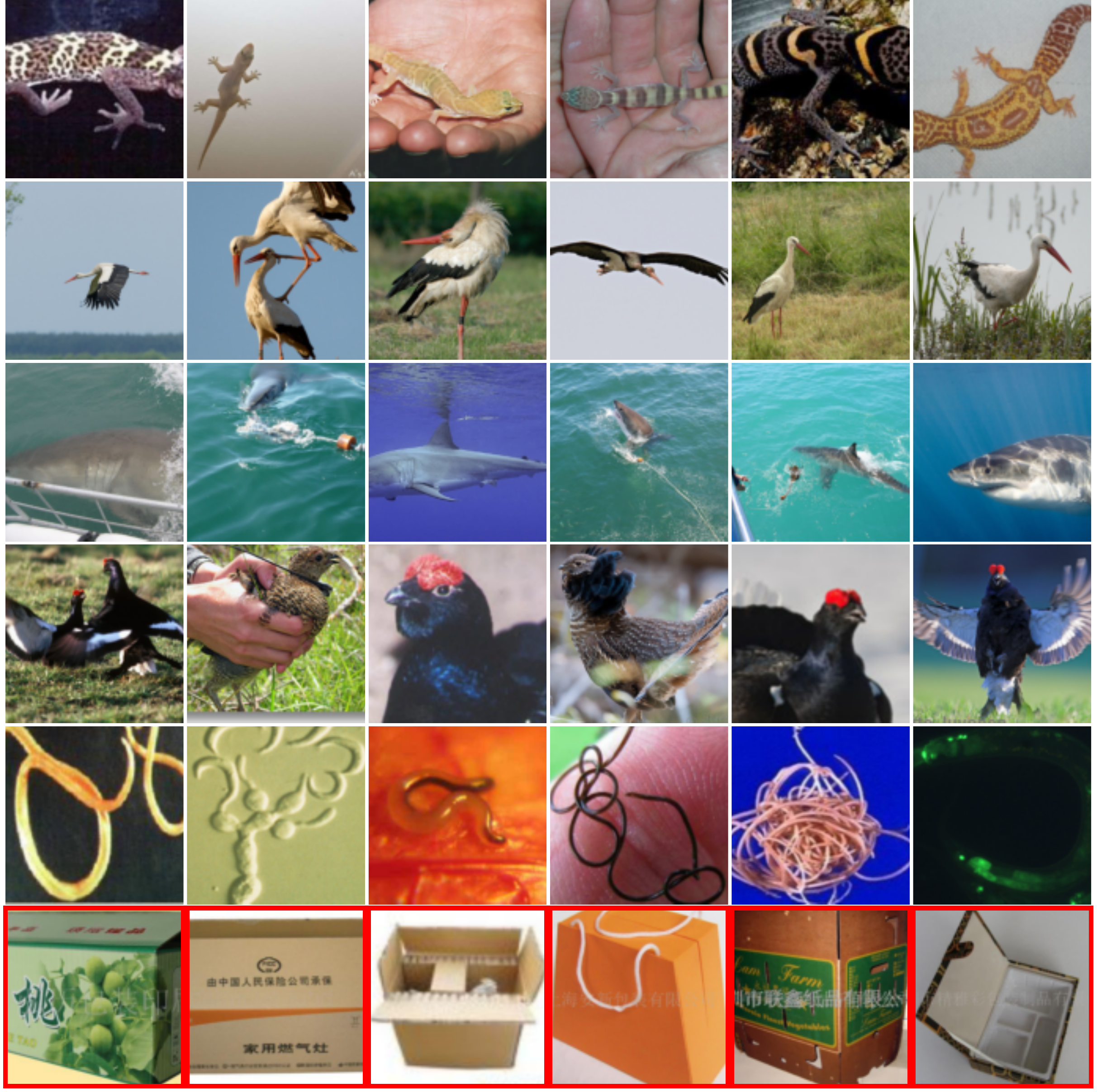}
        \caption{With ``Carton'' class}
        \label{fig:split_imagenet_train_w_carton}
    \end{subfigure}
    \caption{\small \textbf{Training samples in Split ImageNet-100}. Two plots show training samples in a certain task with different bias levels, \ie, 0 \& 6. Each row represents a specific class within the task. The last row of the right plot is the carton class.}
    \label{fig:split_imagenet_train_samples}
\end{figure*}

\begin{figure*}[t!]
    \centering
    \begin{subfigure}[t]{0.4\linewidth}
        \includegraphics[width=0.9\linewidth]{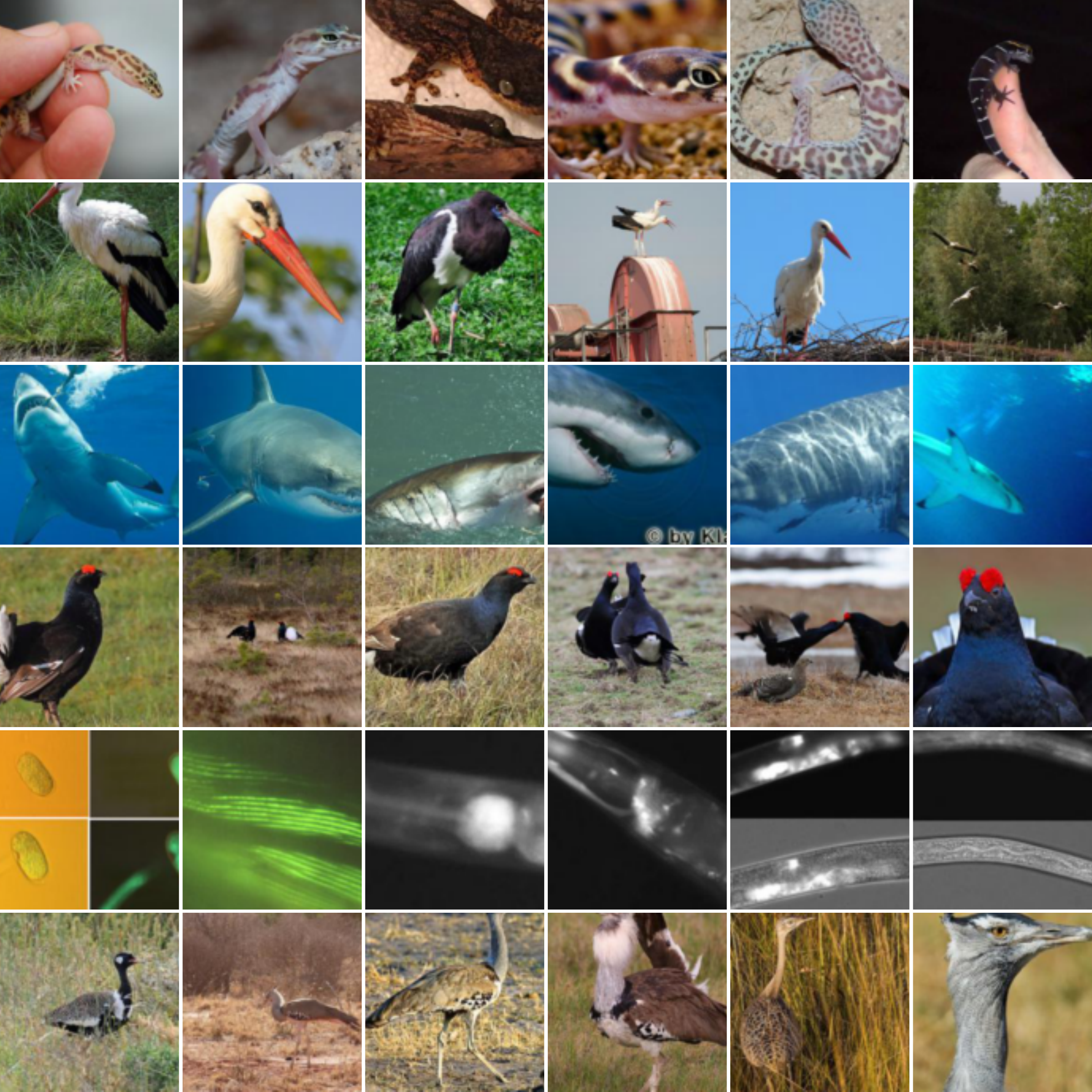}
        \caption{Test samples in a task without watermark}
        \label{fig:split_imagenet_test_wo_water}
    \end{subfigure}
    \begin{subfigure}[t]{0.4\linewidth}
        \includegraphics[width=0.9\linewidth]{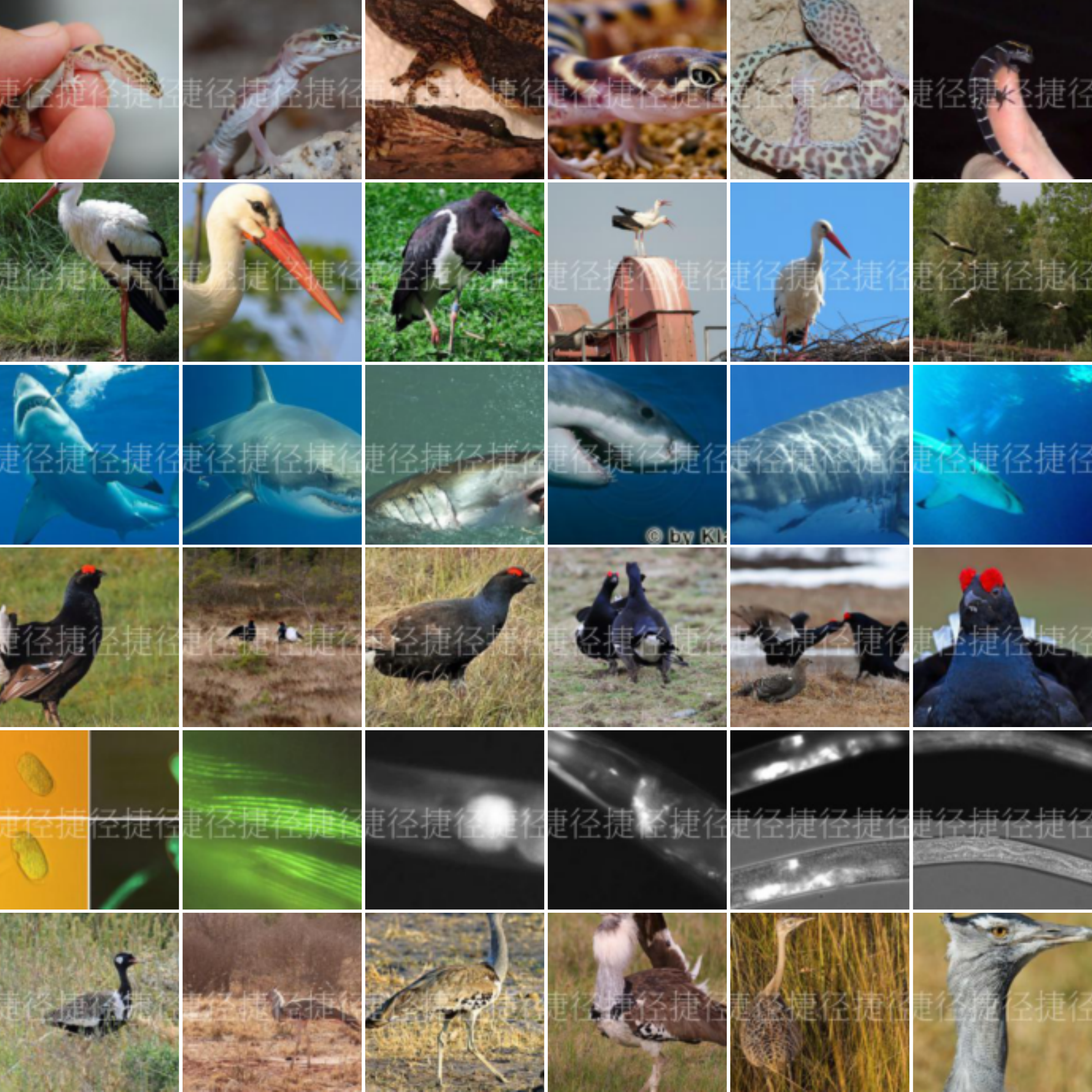}
        \caption{Test samples in a task with watermark}
        \label{fig:split_imagenet_test_w_water}
    \end{subfigure}
    \caption{\small \textbf{Test samples in Split ImageNet-100}. Both sides of samples are used to compute BMR.}
    \label{fig:split_imagenet_test_samples}
\end{figure*}

\noindent\textbf{CelebA}. Domain-IL typically assumes the input distributions vary as the number of tasks increases. To reflect this assumption in CL scenarios, we divided the CelebA dataset into several tasks based on some selected attributes and each task thereby has different facial features. For instance, the two tasks in CelebA$^2$ are defined by whether face images are ``smiling'' or not. Similarly, we utilize three attributes, ``Black Hair'', ``Oval Face'', and ``Mouth slightly open'' for CelebA$^8$. By dividing the data in this way, CL scenarios based on CelebA$^2$ or CelebA$^8$ mimic the distribution shift occurring in real-world applications.

\noindent\textbf{Split ImageNet-100}. To study the impact of the watermark bias in CL scenarios, we replace a randomly selected class with the ``Carton'' class in a certain task. To calculate BMR for the watermark bias, we make watermarked versions of each sample in the original test dataset by the style transfer used in \cite{li2022whac}. Examples of training and test samples of Split ImageNet-100 are illustrated in Fig \ref{fig:split_imagenet_train_samples} and \ref{fig:split_imagenet_test_samples}.

\section{Naive debiasing in CL scenarios}
\label{sec_append:naive_debiasing}
\begin{figure}
    \centering
    \includegraphics[width=0.6\linewidth]{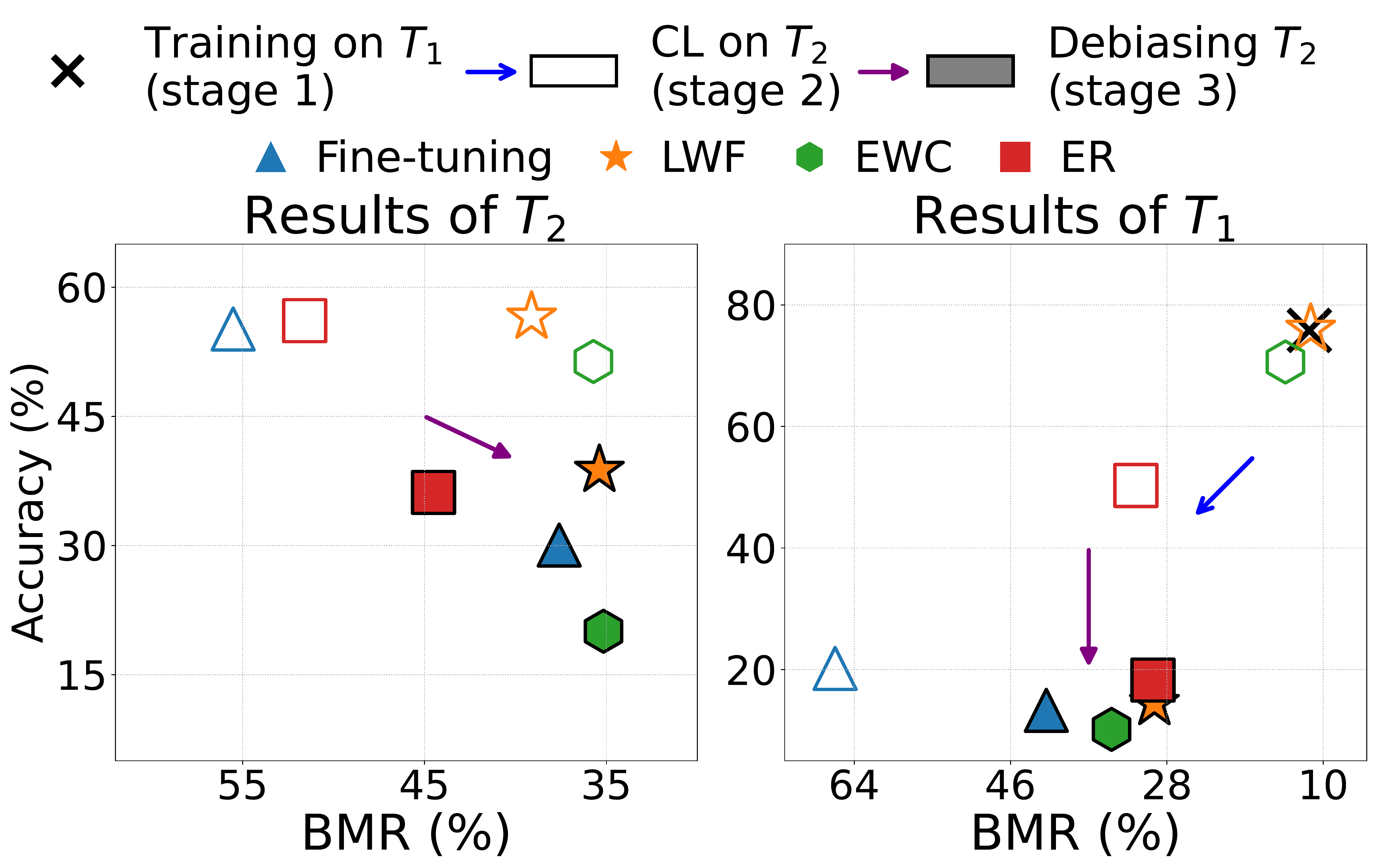}
    \caption{\small \textbf{Naive debiasing with Group DRO on Split CIFAR-100S.} The accuracy and BMR of $T_1$ and $T_2$ are shown for each stage.}
    \label{fig:naive_debiasing_gdro}
\end{figure}

Here, we assume a scenario in which the bias of a model is detected after learning $T_1$ (stage 1) and $T_2$ (stage 2) continually. After that, the model is re-trained using existing debiasing techniques to obtain a debiased model (stage 3). For debiasing, we employ an existing debiasing technique, Group DRO, \cite{groupdro} which minimizes the worst-case group loss.
For this scenario, we set the bias level of $T_1$ and $T_2$ as 0 and 6 respectively. Figure
\ref{fig:naive_debiasing_gdro} show the accuracy and BMR of $T_2$ (left) and $T_1$ (right) at each stage for each baseline. We plotted the results of each baseline with hyperparameters achieving the highest average accuracy of the two tasks.
In the right plot, we observe that some points shift to the bottom left as progressing from stage 1 to stage 2, \ie, forgetting of $T_1$.
From results of the stage 3 in the left plot, we show that BMR of $T_2$ can be reduced by Group DRO and MFD.
However, we also identify that the accuracy of $T_1$ significantly drops. Thus, when debiasing after learning each task as we argued in Section \ref{sec:two_task_studies}, one should consider forgetting issue of the learned tasks at the same time, \ie, it is necessary to develop a novel debiasing method considering the stability for CL.

\section{Additional experimental results}
\label{sec_append:additional_results}
\subsection{Forward and backward transfers of bias for CL with two tasks}
\begin{figure*}[t!]
    \centering
    \begin{subfigure}[t]{0.4\linewidth}
        \includegraphics[width=0.9\linewidth]{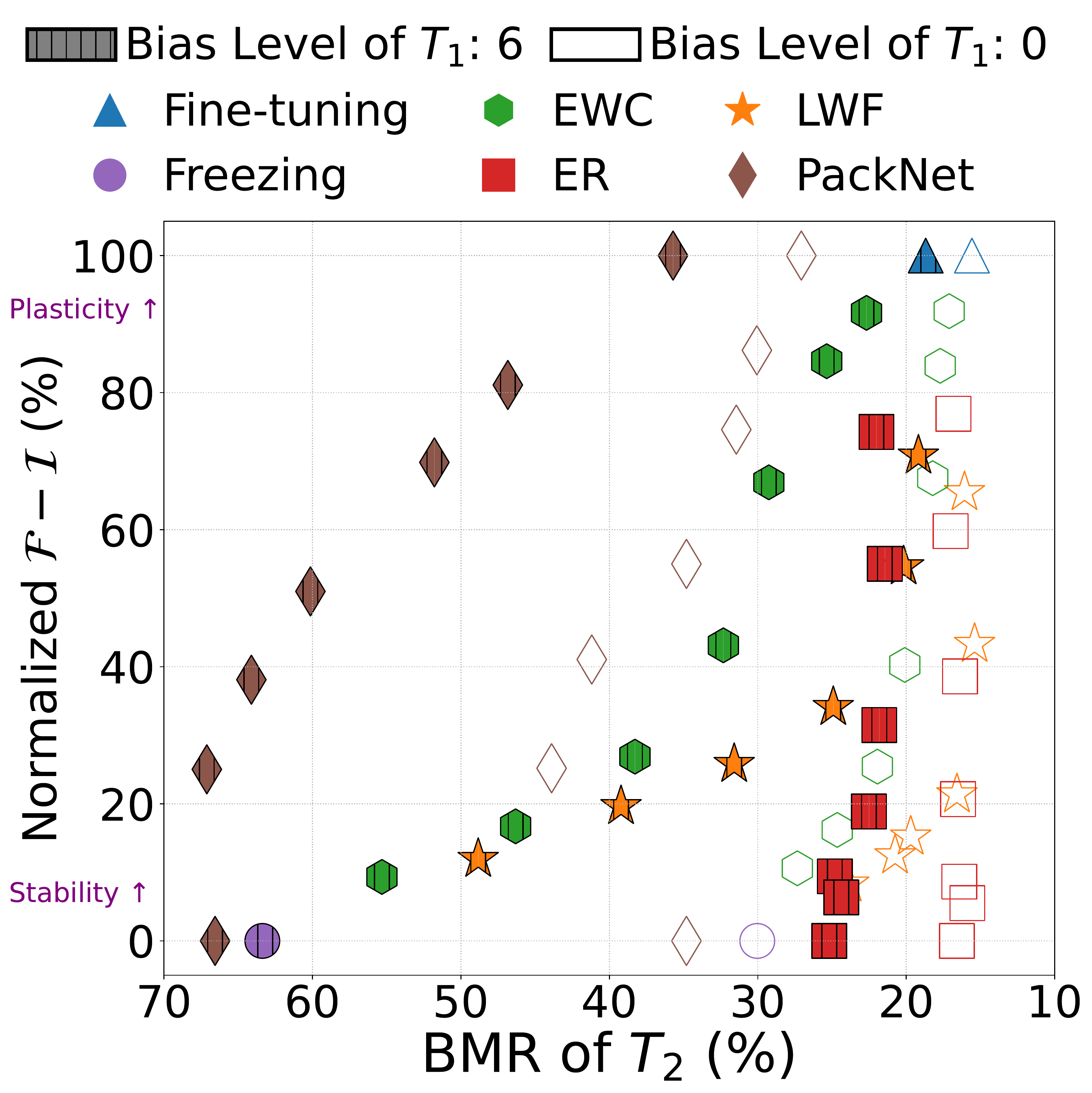}
        \caption{Bias level of $T_2$ : 2}
        \label{fig:two_forward_cifar_level2}
    \end{subfigure}
    \begin{subfigure}[t]{0.4\linewidth}
        \includegraphics[width=0.9\linewidth]{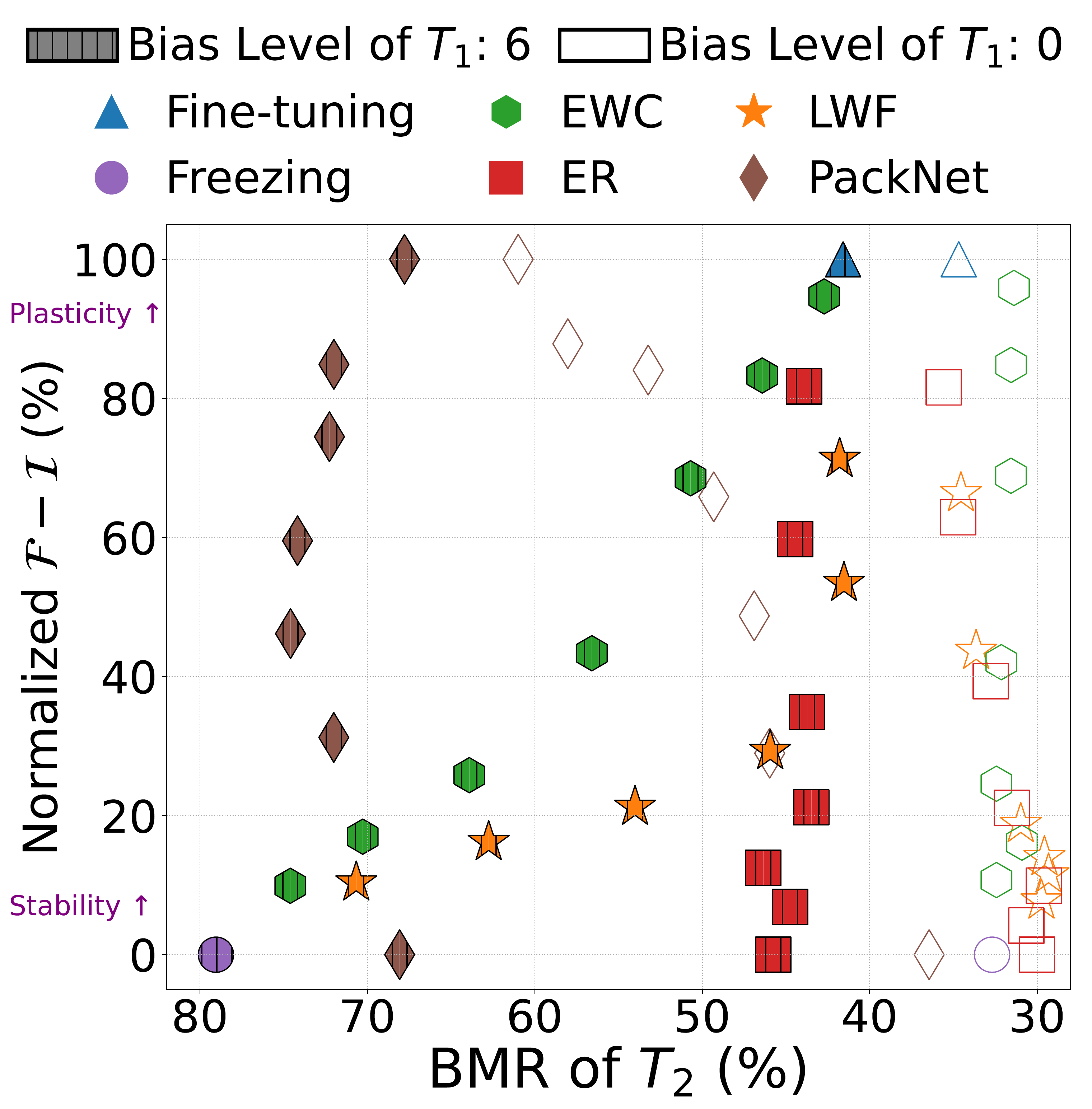}
        \caption{Bias level of $T_2$ : 4}
        \label{fig:two_forward_cifar_level4}
    \end{subfigure}
    \caption{\small{\bf Forward transfer of bias in two tasks-continual learning on Split CIFAR-100S}. }
    \label{fig:two_forward_more_levels}
\end{figure*}

Figure \ref{fig:two_forward_more_levels} displays the forward transfer of the color bias for two task-CLs on Split CIFAR-100S. In each plot in the figure, the bias levels of $T_2$ are fixed to 2 or 4, respectively, and BMR of $T_1$ is reported for each CL method and hyperparameter. It is apparent from the figure that the difference in BMR between colored and uncolored points becomes more pronounced with the bias transfer, as compared to results obtained when the bias level of $T_2$ is fixed to 0. This would be because previously learned biases of a CL model tend to facilitate learning of the dataset bias of the current task more. 

\begin{figure*}[t!]
    \centering
    \begin{subfigure}[t]{0.4\linewidth}
        \includegraphics[width=0.9\linewidth]{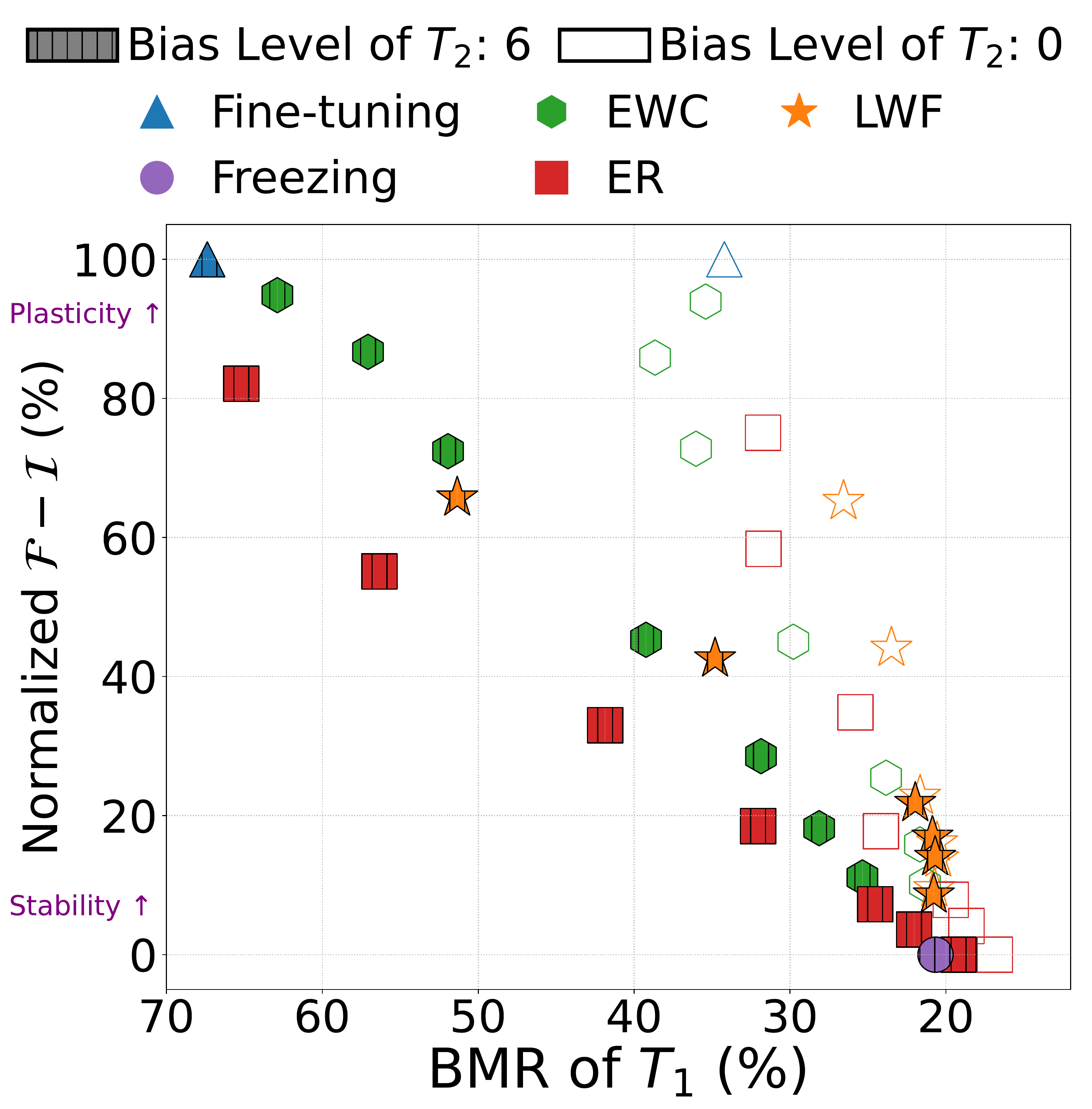}
        \caption{Bias level of $T_1$ : 2}
        \label{fig:two_backward_cifar_level2}
    \end{subfigure}
    \begin{subfigure}[t]{0.4\linewidth}
        \includegraphics[width=0.9\linewidth]{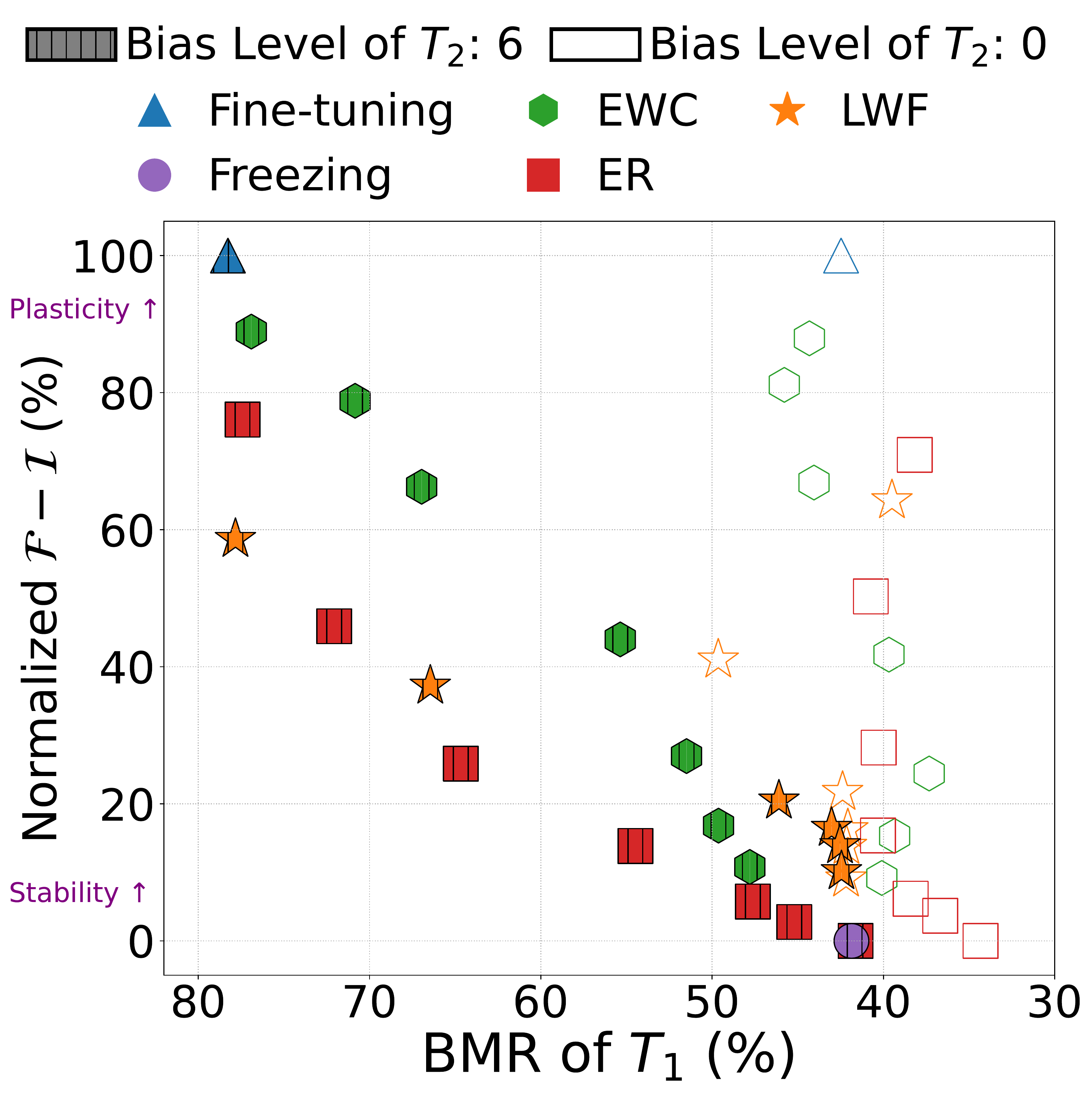}
        \caption{Bias level of $T_1$ : 4}
        \label{fig:two_backward_cifar_level4}
    \end{subfigure}
    \caption{\small{\bf Backward transfer of bias in two tasks-continual learning on Split CIFAR-100S}.}
    \label{fig:two_backward_more_levels}
\end{figure*}

Similarly, in Figure \ref{fig:two_backward_more_levels}, we present the outcomes of two-task experiments for analysis of the backward transfer of bias. As in Figure \ref{fig:two_backward_more_levels}, the results show that if $T_1$ already contains a dataset bias, the effect of the backward transfer of the bias from $T_2$ is more pronounced. 

\subsection{Feature representation analysis for backward transfer of bias}
\begin{figure}
    \centering
    \includegraphics[width=0.4\linewidth]{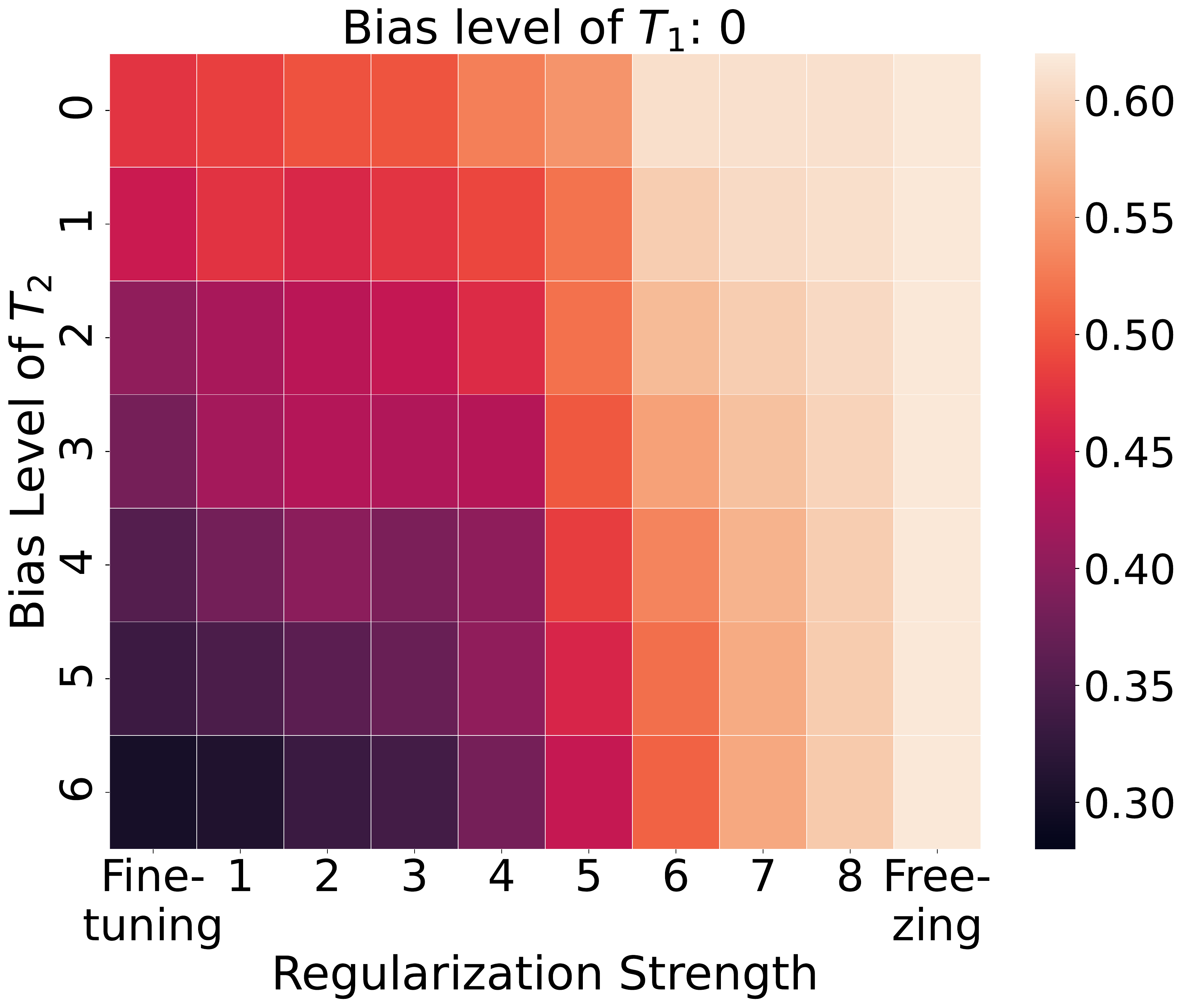}
    \caption{\small {\bf CKA on Split CIFAR-100S.} The CKA values between color images and grayscale images in $T_1$ are shown. Each value is calculated after learning $T_2$ by EWC.}
    \label{fig:cka_backward}
\end{figure}

We exhibit the results of CKA analysis for the backward transfer on Split CIFAR-100S. From Figure \ref{fig:cka_backward}, we observe a clear trend indicating the CKA value decreases as the regularization strength decreases and the bias level of $T_2$ increases. This again suggests that when a CL method focuses on learning a biased current task, \ie, plasticity, the backward transfer of bias by a CL method becomes more obvious. 

\subsection{Experimental results for accuracy with a longer sequence of tasks}
\begin{figure}
    \centering
    \begin{subfigure}{0.45\linewidth}
        \centering
        \includegraphics[width=0.9\linewidth]{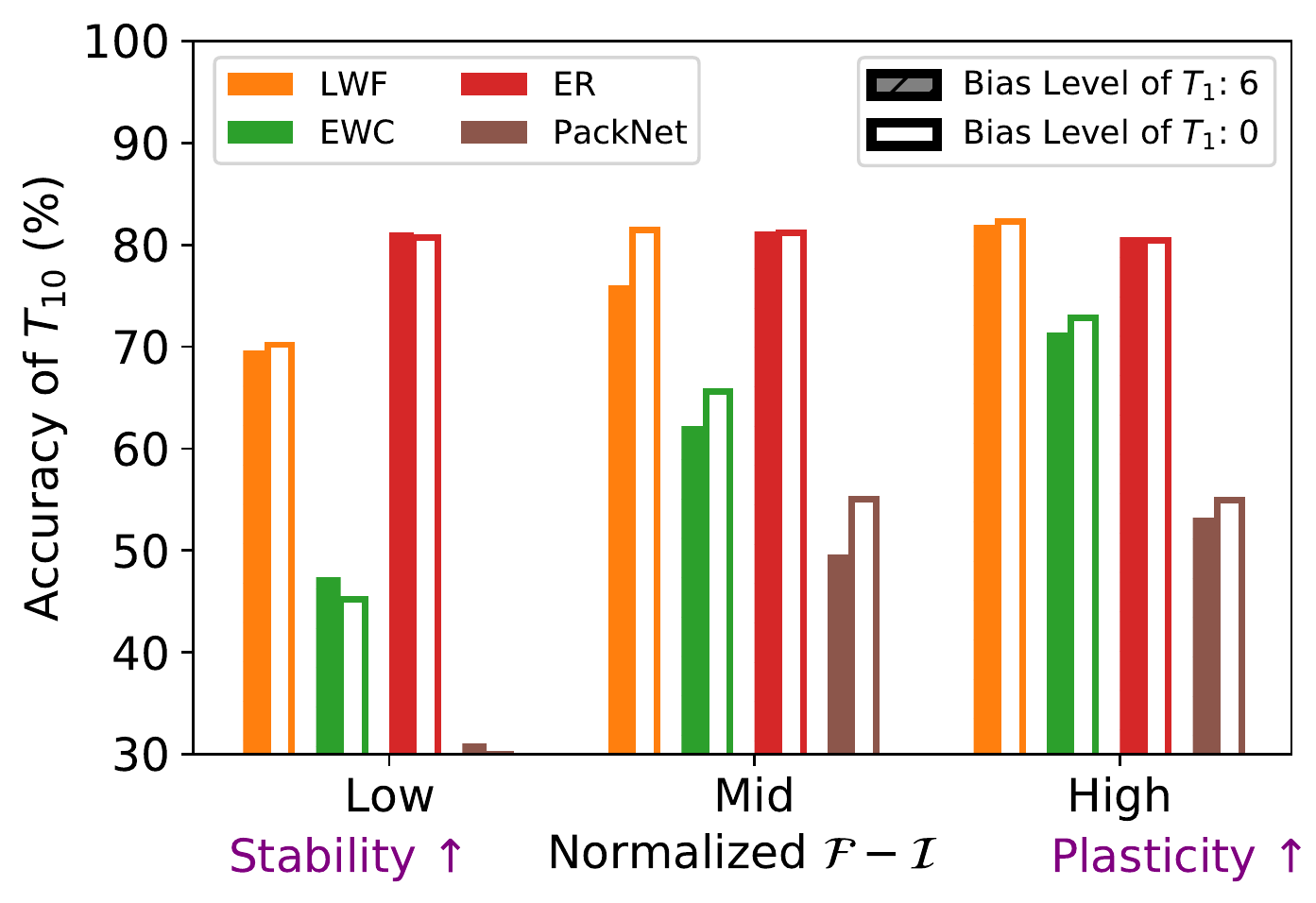}    
    \end{subfigure}
    \begin{subfigure}{0.45\linewidth}
        \centering
        \includegraphics[width=0.9\linewidth]{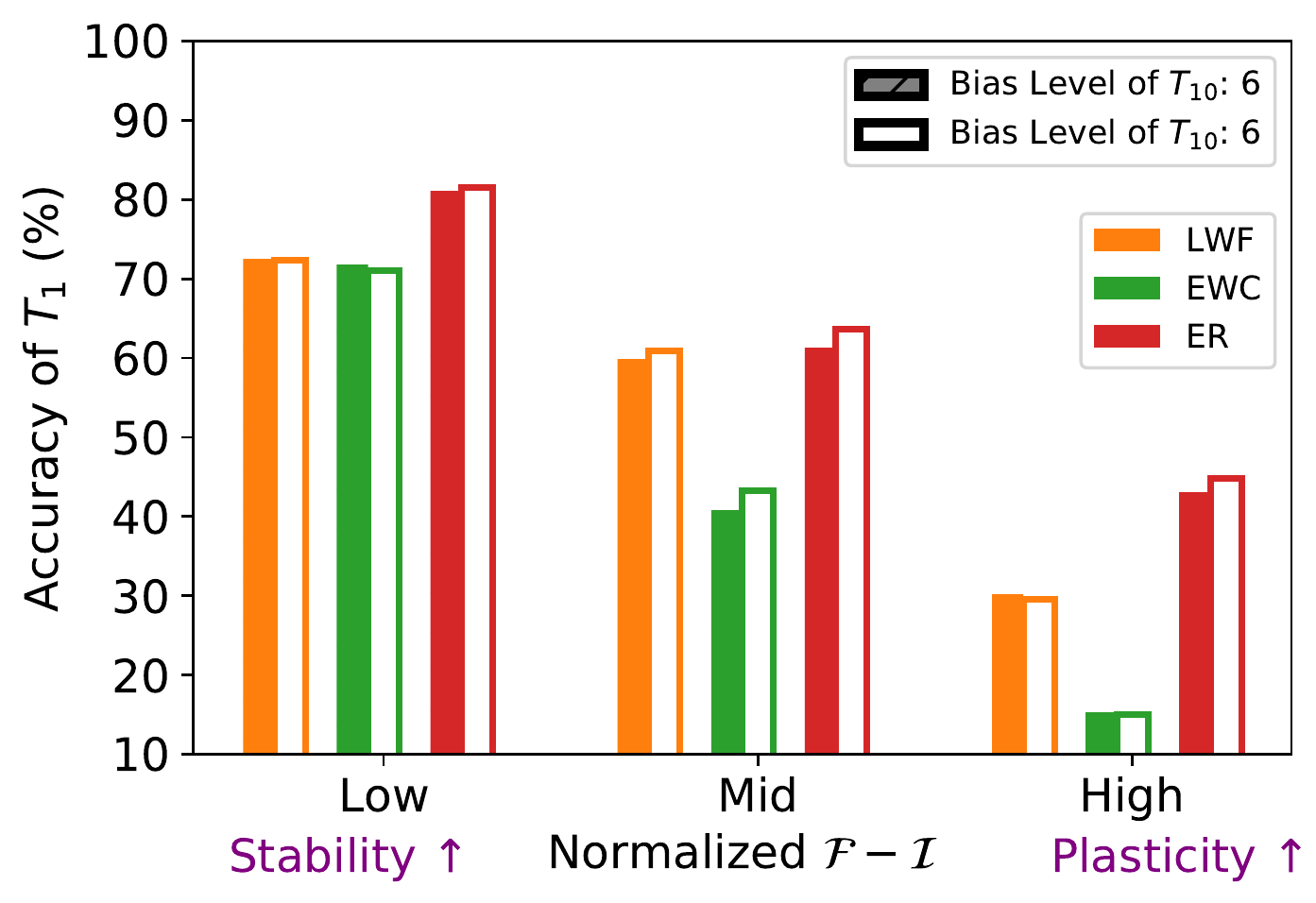}    
    \end{subfigure}    
    \caption{\small {\bf Accuracy in longer sequences of Split CIFAR-100S.} The experimental settings in two plots are the same as in Figure 5(a) and 5(b), respectively.}
    \label{fig:long_cifar_acc}
\end{figure}

\begin{figure}
    \centering
    \includegraphics[width=0.4\linewidth]{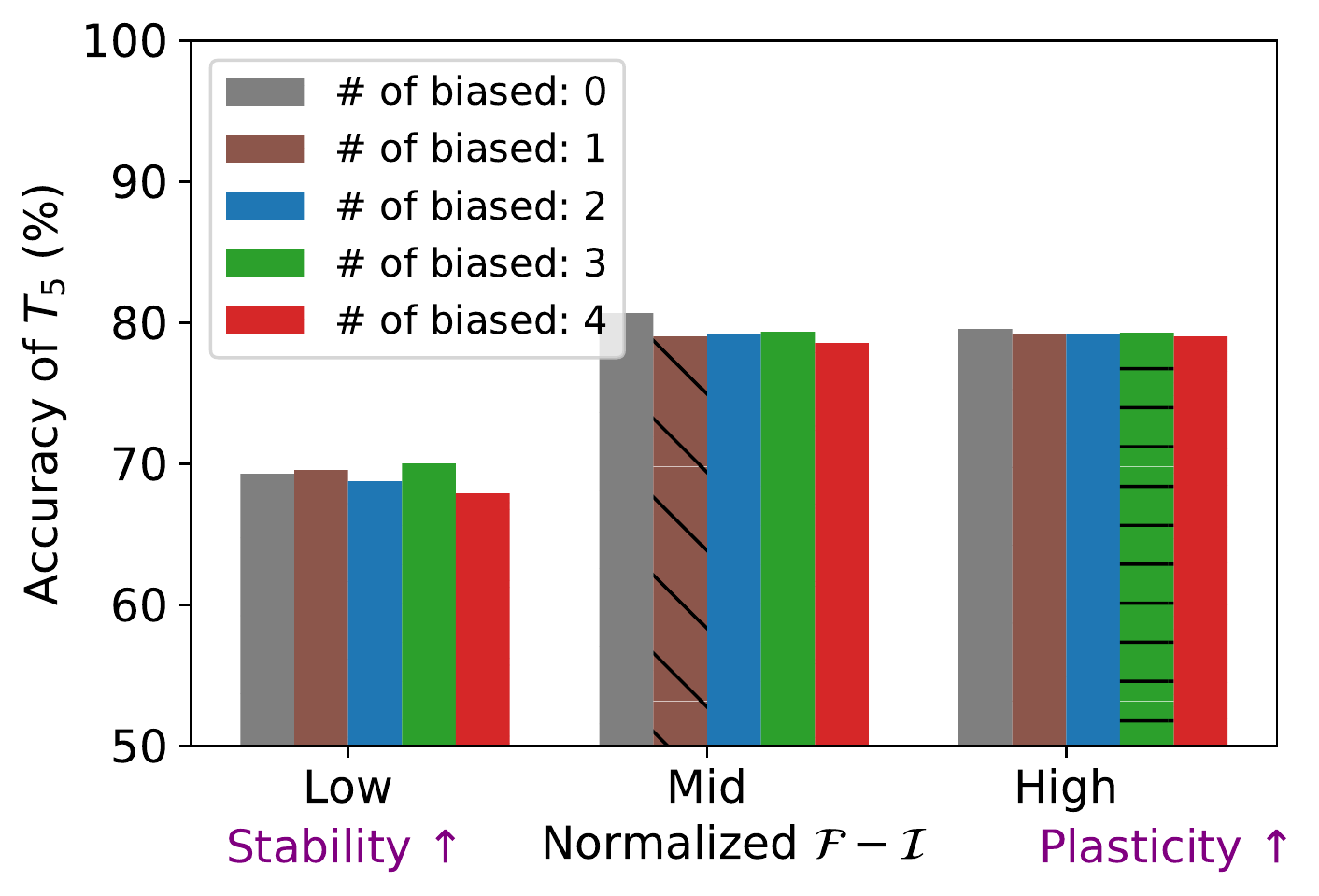}
    \caption{\small{\bf Accuracy of $T_5$ depending on the number of biased tasks.} The experimental settings are the same as in Figure 6.}
    \label{fig:accumul_acc}
\end{figure}

Figure \ref{fig:long_cifar_acc} and \ref{fig:accumul_acc} show accuracies corresponding to each of the results in Figure 5 and 6 in the manuscript. The figures demonstrate that the accuracies in the same intervals are roughly the same, so we conclude that the gaps of BMR shown in Figure 5 and 6 are due to the bias transfers, not the accuracy gaps.

\begin{table}[]
\centering
\caption{\small\textbf{Mis-classified ratio of $T_2$ test data when watermark is added either after learning ``carton'' in $T_1$ or not.} We reported the results with memory capacity which is 100\% of $T_1$ data to consider the stability.}
\resizebox{0.5\linewidth}{!}{
\begin{tabular}{ccc}
\toprule
\multirow{2}{*}{Prediction} & \multicolumn{2}{c}{Mis-classified ratio (\%)}              \\ 
                             & \multicolumn{1}{c}{$T_1$ with ``carton''} & $T_1$ without ``carton'' \\ \midrule
Old class                  & \multicolumn{1}{c}{4.52}             & 7.78                \\ 
New class                  & \multicolumn{1}{c}{7.74}             & 6                \\ 
``Carton'' class (old)            & \multicolumn{1}{c}{4.02}             & -                   \\ \midrule 
Total (BMR)             &      16.28   &    13.78           \\ \bottomrule
\end{tabular}
}
\label{tab:forward_image_mis}
\end{table}

\begin{table}[]
\centering
\caption{\small \textbf{Mis-classified ratio of $T_1$ test data when watermark is added either after learning ``carton'' in $T_2$ or not.} We reported the results with memory capacity which is 10\% of $T_1$ data to consider the plasticity.}
\resizebox{0.5\linewidth}{!}{
\begin{tabular}{ccc}
\toprule
\multirow{2}{*}{Prediction} & \multicolumn{2}{c}{Mis-classified ratio (\%)}              \\ 
                             & \multicolumn{1}{c}{$T_2$ with ``carton''} & $T_2$ without ``carton'' \\ \midrule
Old class                  & \multicolumn{1}{c}{2.06}             & 1.38              \\ 
New class                  & \multicolumn{1}{c}{21.61}            & 28.06               \\ 
``Carton'' class (new)            & \multicolumn{1}{c}{5.69}            & -                   \\ \midrule 
Total (BMR)             &    29.36     &  29.44            \\ \bottomrule
\end{tabular}
}
\label{tab:backward_image_mis}
\end{table}

\section{Discussions about results for backward transfer on Split ImageNet-100}
\label{sec_append:discussion}
In this section, we analyze the predictions of a models in order to investigate why the backward transfer is not observed in Class-IL scenarios with two-tasks on Split ImageNet-100. 
Table \ref{tab:forward_image_mis} (resp. Table \ref{tab:backward_image_mis}) represents the CL scenarios for the forward (resp. backward) transfer of bias,  which evaluates the bias for $T_2$ (resp. $T_1$) by varying the bias level of $T_1$ (resp. $T_2$), \ie, whether the ``carton'' class is contained in the task or not. We employ ER in our experiments and set the memory size as 1 in Table \ref{tab:forward_image_mis} or 0.1 in Table \ref{tab:backward_image_mis}, to make forward and backward transfer of bias significantly occurs by focusing more on stability or stability, respectively. 
To analyze BMR values in more detail, we divided the cases of misclassification  of bias-conflicted samples into three categories; a CL model falsely predicts for 1) one of the old classes (except the carton), 2) one of the new classes (except the carton), and 3) the ``carton'' class. 

The following are our observations from the tables.
First, we observe that BMRs of $T_1$ in Table \ref{tab:backward_image_mis} show almost no difference regardless of whether a CL model learns the ``carton'' class under high plasticity, while BMR of $T_2$ increases in Table \ref{tab:forward_image_mis} when $T_1$ contains the ``carton'' class. Namely, in terms of BMR, it looks like the backward transfer of the watermark bias does not occur. However, when we look at the categorized results, we can derive different trends, \ie, the backward transfer still occurs. In Table \ref{tab:backward_image_mis}, when $T_2$ contains the ``carton'' class, some samples with the watermark are misclassified into the ``carton'' class, \ie, the CL model possesses watermark bias in $T_1$ due to backward transfer. We then argue that the main cause of similar BMRs in Table \ref{tab:backward_image_mis} would be the \textit{old-new bias}, an inherent issue of Class-IL, which indicates biased predictions towards new classes. Indeed, when the watermark is injected into samples in $T_1$ which belongs to old classes, we see a disproportionately high ratio of incorrectly predicted samples for the new classes in Table \ref{tab:backward_image_mis}. On the other hand, we observe that the ratios for old and new classes are relatively similar in Table \ref{tab:forward_image_mis}, since the watermark bias is injected to samples in $T_2$ which belongs to new classes. Thus, we can infer that the old-new bias can make the model predictions for samples in $T_1$ vulnerable to watermark bias and easily shift to new classes in $T_2$, even if the model did not explicitly learn the ``carton'' class. This explains high misclassified ratios for new class regardless of the dataset bias of $T_2$ and similar BMRs in Table \ref{tab:backward_image_mis} although the backward transfer of bias occurs. 

\end{document}